\documentclass[acmtog,dvipsnames,screen,nonacm]{acmart}  % ArXiv version

\acmSubmissionID{2184}

\usepackage{booktabs} % For formal tables
\usepackage[abs]{overpic}
\usepackage{algorithmicx,algpseudocode}
\usepackage{graphicx}
\usepackage{multirow}
\usepackage{pifont}
\usepackage{dsfont}
\usepackage{soul}
\usepackage{enumitem}
\usepackage[table]{xcolor} 
\usepackage{subcaption}
\usepackage{tikz}

\usepackage{soul}
\usepackage[dvipsnames]{xcolor}
\usepackage{colortbl}

\newcommand{\myparagraph}[1]{\vspace{0.5em}\noindent\textbf{#1}}
\newcommand{\Tref}[1]{Table~\ref{#1}}

\newcommand{\fref}[1]{Fig.~\ref{#1}}
\newcommand{\Fref}[1]{Figure~\ref{#1}}

\def\eg{\emph{e.g.}}

\definecolor{SceneBlue}{rgb}{0.7031,    0.7812,    0.8632}
\definecolor{SceneRed}{rgb}{0.8867,    0.6171,    0.5781}
\definecolor{SceneOrange}{rgb}{0.9296,    0.8125,    0.5859}
\definecolor{ScenePurple}{rgb}{0.7265,    0.6289,    0.7773}
\definecolor{SceneGreen}{rgb}{0.6562,    0.7656,    0.5937}
\definecolor{LightGreen}{rgb}{0.835,    0.835,    0.835}

\usepackage{amsmath,amsfonts,bm}

\def\eqref#1{equation~\ref{#1}}
\def\1{\bm{1}}

\DeclareMathAlphabet{\mathsfit}{\encodingdefault}{\sfdefault}{m}{sl}
\SetMathAlphabet{\mathsfit}{bold}{\encodingdefault}{\sfdefault}{bx}{n}

\DeclareMathOperator*{\argmin}{arg\,min}

\usepackage[ruled]{algorithm2e} % For algorithms

\SetAlFnt{\small}
\SetAlCapFnt{\small}
\SetAlCapNameFnt{\small}
\SetAlCapHSkip{0pt}

\acmJournal{TOG}
\newcolumntype{a}{>{\columncolor[HTML]{EFEFEF}}c}
\newcolumntype{b}{>{\columncolor{yellow}}c}

\begin{document}
% Title portion
% \title{Unified Surface Keypoint Representation for Human-Object Interaction Generation}
\title{Surface Keypoint Representation for Multi-Object and Articulated Human-Object Interaction Generation}

% DO NOT ENTER AUTHOR INFORMATION FOR ANONYMOUS TECHNICAL PAPER SUBMISSIONS TO SIGGRAPH 2019!
\author{Xiaogang Peng}
% \orcid{1234-5678-9012-3456}
\affiliation{%
 \institution{Northeastern University}
 \streetaddress{360 Huntington Ave}
 \city{Boston}
 \state{MA}
 \postcode{02115}
 \country{USA}}
\email{peng.xiaog@northeastern.edu}

\author{Zeyu Han}
\affiliation{%
 \institution{Northeastern University}
 % \streetaddress{360 Huntington Ave}
 % \city{Boston}
 % \state{MA}
 % \postcode{02115}
 \country{USA}
 }
\email{han.zeyu@northeastern.edu}

\author{Zichong Meng}
\affiliation{%
 \institution{Northeastern University}
 % \streetaddress{360 Huntington Ave}
 % \city{Boston}
 % \state{MA}
 % \postcode{02115}
 \country{USA}
 }
\email{meng.zic@northeastern.edu}

\author{Yiming Xie}
\affiliation{%
 \institution{Northeastern University}
 % \streetaddress{360 Huntington Ave}
 % \city{Boston}
 % \state{MA}
 % \postcode{02115}
 \country{USA}
 }
\email{ymxyimingxie@gmail.com}

\author{Jihua Zhu}
\affiliation{%
 \institution{Xi'an Jiaotong University}
  \country{China}
 }
\email{zhujh@xjtu.edu.cn}

\author{Gang Hua}
\affiliation{%
 \institution{Amazon}
  \country{USA}
 }
\email{ganghua@gmail.com}

\author{Huaizu Jiang}
\affiliation{%
 \institution{Northeastern University}
 % \streetaddress{360 Huntington Ave}
 % \city{Boston}
 % \state{MA}
 % \postcode{02115}
 \country{USA}
 }
\email{h.jiang@northeastern.edu}

\begin{abstract}
Daily activities require humans to coordinate whole-body motion with the motion of surrounding objects.
Despite recent progress in human-object interaction (HOI) generation, most existing methods assume interactions with a single rigid object and do not extend well to scenarios involving a variable number of objects or articulated objects with diverse joint mechanisms.
We propose surface keypoint trajectories as an object motion representation: for each rigid component, whether a standalone object or one part of an articulated assembly, we track a small set of non-collinear surface points over time.
This representation handles multi-object coordination and diverse articulation mechanisms directly from point dynamics without requiring explicit joint-type specification.
To model when and where each body region contacts each object, we introduce a spatio-temporal contact distance field that extends distance-based contact modeling to whole-body, multi-object, and articulated settings.
We factorize HOI generation into three stages: generating object motions from text or waypoints, predicting the contact distance field, and synthesizing whole-body motion with contact-guided optimization.
Experiments on ParaHome, HIMO, ARCTIC, and OMOMO demonstrate better or comparable performance to existing methods across single-object, multi-object, and articulated interaction settings.
\end{abstract}


%
% End generated code
%


\begin{teaserfigure}
    \includegraphics[width=\linewidth]{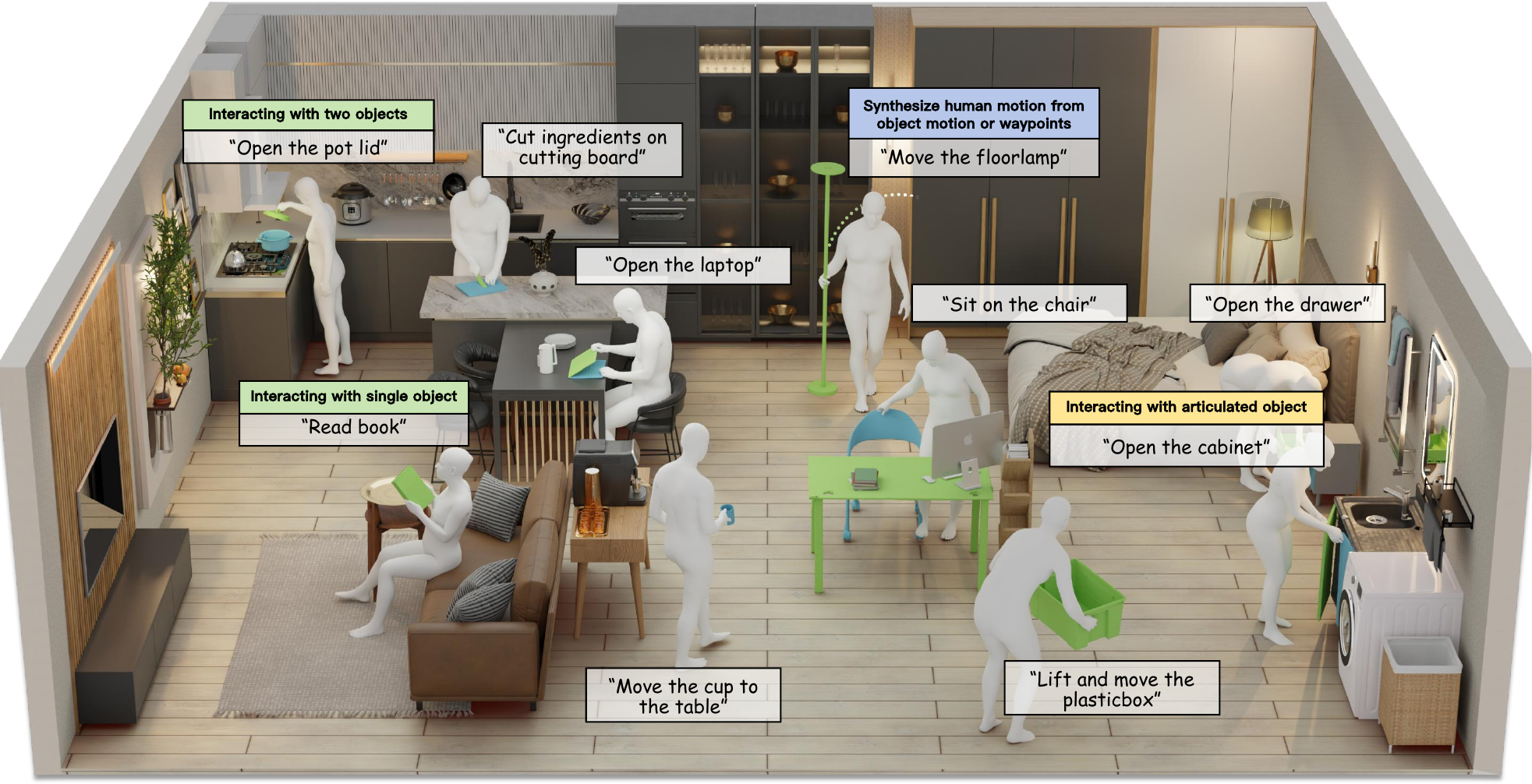}
    % \caption{\textbf{Autonomous \acs{hsi} synthesis}. Our proposed method generates realistic character motion in 3D scenes based on a single textual instruction and goal location, incorporating seamless transitions between locomotion and \acs{hoi} autonomously.}
    \label{fig:teaser}
    % \vspace{-2.5em}
    \caption{Our approach enables: (a) whole-body interaction of a variable number of rigid objects from text prompts; (b) interaction with articulated objects exhibiting diverse joint dynamics (e.g., laptop, drawers, and knobs); and (c) text-driven whole-body motion synthesis guided by sparse object waypoints or full object motion trajectories.}
\end{teaserfigure}

\maketitle
\section{Introduction}
\label{sec:intro}

Everyday activities such as cooking, cleaning, and furniture assembly require humans to coordinate whole-body motion with the movement of surrounding objects.
Synthesizing such human-object interactions (HOIs) is a fundamental challenge in computer graphics and vision, with broad applications in video games, virtual and augmented reality, embodied AI, and robotics.
The goal is to generate natural and contextually appropriate motions for both the human body and the objects being manipulated.

Previous works have made notable progress in HOI generation conditioned on various input modalities, including textual descriptions~\cite{diller2023cghoi, peng2023hoi}, sparse object waypoints/trajectories~\cite{li2023controllable, li2023object}, object geometry and target grasps~\cite{taheri2021goal, wu2022saga}, past motion history~\cite{xu2023interdiff}, and videos~\cite{li2026zerohsi}.
However, most existing methods face two key limitations when applied to more complex real-world interactions.

\textbf{(i) Multi-object and articulated interaction generation.}
Most existing methods focus on interactions with a single rigid object.
However, multi-object interactions introduce additional complexity, as the number and combination of objects involved varies across activities, requiring the generation model to handle flexible object configurations.
Moreover, the dominant object representation, global $SE(3)$ transformations~\cite{li2023object, li2023controllable, diller2023cghoi, peng2023hoi}, inherently assumes rigid-body motion.
% and does not naturally extend to articulated objects.
Although per-part $SE(3)$ transformations can describe articulated objects~\cite{pi2025coda, zhang2025bimart}, they require explicit joint-type specification, with separate handling needed for each mechanism (revolute, prismatic, screw).
% \hj{Do existing approaches require different models for different joint types? Is ``generalization'' supported by our approach?} \xg{Existing methods primarily handle revolute joint types and separate models would be required for other joint types. Our method can be more general for different joint types.}
While recent benchmarks such as HIMO~\cite{lv2024himo} and ParaHome~\cite{kim2025parahome} have begun to capture multi-object and articulated interactions, the generation methods trained on them still require pre-specifying the number of objects~\cite{lv2024himo}.
% or rely on per-part $SE(3)$ plus explicit joint-type labels.
Furthermore, existing methods either generate human and object motions jointly~\cite{diller2023cghoi, peng2023hoi}, which struggles as the number of objects grows, or require object trajectories as input~\cite{li2023object}, limiting their applicability.
As a result, generation methods that handle a variable number of objects and diverse articulation mechanisms, while also producing object motions directly from text, remain limited.

\textbf{(ii) Contact modeling.}
Accurately capturing when and where each body part contacts each object is essential for physically plausible interactions, especially when multiple objects are involved.
Yet existing contact representations are either binary labels indicating presence or absence of contact~\cite{diller2023cghoi, peng2023hoi}, which lack fine-grained spatial information; static affordances that do not evolve over time~\cite{liu2023contactgen, cseke2025pico}; or hand-centric correspondences that do not extend to whole-body interactions~\cite{li2023controllable, pi2025coda}.
ManipNet~\cite{he2021manipnet} introduces distance-based spatial sensors between hands and objects, but these are local to the hand and do not generalize to whole-body, multi-object settings.
% ROG~\cite{xue2025guiding} proposes an Interactive Distance Field between skeleton joints and object boundary keypoints, but it is limited to single-object interactions and does not model temporal evolution. \xg{ROG captures temporal dynamics, but it represents contact context on the object surface too sparsely.}
ROG~\cite{xue2025guiding} constructs an Interactive Distance Field for HOI generation, but handles only single-object interactions, measures proximity from sparse interior skeletal joints rather than the body surface, and uses a limited number of object keypoints.
None of these approaches provides fine-grained, temporally dynamic correspondence between whole-body markers and multiple object surfaces, the kind of signal needed to guide generation when different body parts engage with different objects at different times.

\begin{figure}[t]
    \centering
    \includegraphics[width=1.0\linewidth]{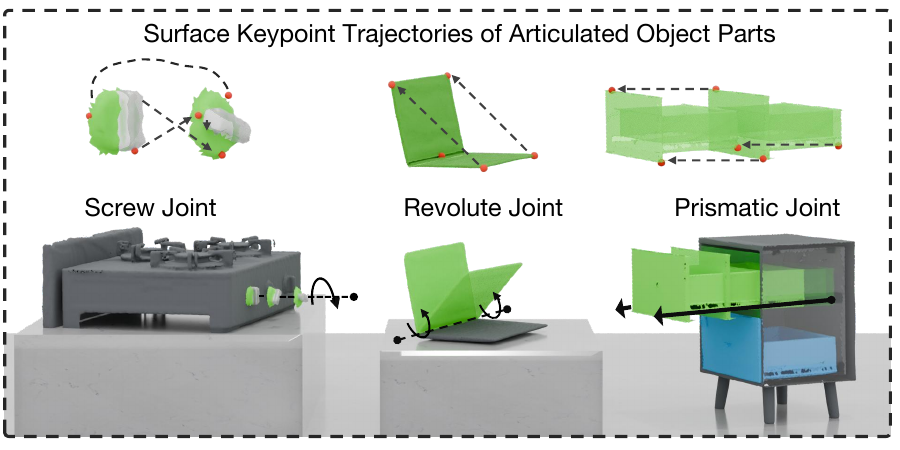}
    \vspace{-2.5em}
    \caption{\textbf{Surface keypoint trajectories can represent diverse articulated motion.}
For each articulated object, we represent every component using a small set of surface keypoints. Different articulation mechanisms induce distinct trajectory patterns: screw joints produce coupled rotation and translation, revolute joints produce rotation around a fixed axis, and prismatic joints produce linear translation. 
% By modeling the motion of each part through its surface keypoint trajectories, our representation captures these diverse articulated motions without requiring explicit joint-type specification.
}
    \label{fig:articulated}
    \vspace{-1.7em}
\end{figure}

To address limitation (i), we propose \emph{surface keypoint trajectories} as an object motion representation.
For each rigid component, whether a standalone object or one part of an articulated assembly, we sample a small set of non-collinear surface points from its mesh and track their 3D positions over time.
A key property is that three non-collinear points suffice to uniquely determine a rigid transformation via the Kabsch algorithm~\cite{Kabsch:a12999}, so the representation is lossless for rigid motion while operating entirely in Euclidean space.
Different articulation mechanisms (revolute, prismatic, screw, as illustrated in Fig.~\ref{fig:articulated}) each produce distinctive point trajectory patterns that a generative model can learn directly from data, without requiring explicit joint-type specification.
To handle a variable number of rigid components across different interactions, each component occupies a fixed-size slot that is zero-padded when unused and masked during training, allowing a single model to handle one object, multiple objects, or multi-part articulated assemblies without architectural changes.

To address limitation (ii), following prior work~\cite{wu2022saga, zhang2025bimart}, we represent the human body using surface markers on SMPL-X~\cite{SMPL-X:2019}.
This allows us to introduce a \emph{spatio-temporal contact distance field} that captures, for every time step, the proximity between each body marker and each object surface point (1 indicates contact and 0 not).
Unlike the binary, static, or hand-centric contact representations discussed above, our field provides whole-body coverage across multiple objects and their articulated parts. 
While ROG~\cite{xue2025guiding} introduces a distance field for single-object interactions, our formulation extends this concept to multi-object and articulated settings, computes distances from body surface markers rather than interior skeletal joints where contact physically occurs, and uses finer sampling on both the human and object sides. 
By providing fine-grained marker-to-surface correspondences at every time step, the field specifies which body part should contact which object region and when, enabling precise contact-guided optimization during body motion synthesis.

The contact distance field serves as the key intermediate representation in our factorized generation pipeline: given a text prompt, we first generate object keypoint trajectories (Stage~I), then predict the contact distance field conditioned on the generated object motions (Stage~II), and finally synthesize whole-body motion with contact-guided optimization using the predicted field (Stage~III).
This factorization decouples object dynamics, contact prediction, and body synthesis into well-defined sub-problems, allowing each stage to focus on a specific aspect of the interaction while remaining tractable for multi-object and articulated scenarios.
We evaluate our method on four benchmarks that collectively span single-object, multi-object, and articulated interaction settings: ParaHome~\cite{kim2025parahome}, ARCTIC~\cite{fan2023arctic}, HIMO~\cite{lv2024himo}, and OMOMO~\cite{li2023object}.
% We additionally evaluate on the  benchmarks.
Our approach achieves state-of-the-art performance, consistently outperforming existing methods in both motion quality and interaction accuracy.

To summarize, our contributions are as follows.

\begin{itemize}[itemsep=0pt,topsep=0pt,leftmargin=18pt]

    \item We propose surface keypoint trajectories as an object motion representation that enables a single generative model to handle a variable number of rigid and articulated objects without explicit kinematic specification.
    We show this representation outperforms $SE(3)$-based alternatives and naturally supports multi-part composition. % through a zero-padded slot design.

    % \item We introduce a spatio-temporal contact distance field that captures fine-grained, temporally evolving body-object correspondences, providing richer intermediate supervision than binary contact labels or static affordance maps.
    % We show it outperforms alternative contact representations and is essential for multi-object interactions where different body parts engage with different objects at different times.

    \item We introduce a spatio-temporal contact distance field that extends distance-based contact modeling to whole-body, multi-object, and articulated settings, capturing fine-grained, temporally evolving body-object correspondences. Predicted as a standalone intermediate representation, it provides richer supervision than binary contact labels and outperforms alternative contact formulations.

    \item Experiments on ParaHome, OMOMO, ARCTIC, and HIMO demonstrate state-of-the-art performance in both motion quality and interaction accuracy, across single-object, multi-object, and articulated interaction settings.

\end{itemize}

%%% why diffusion models
\section{Related Works}
\label{sec:related_work}

\myparagraph{Representations for Human and Object Motion in HOI Generation.}
Most existing HOI generation methods represent the interaction state by combining a skeleton-based human representation (e.g., joint positions~\cite{Guo_2022_CVPR,meng2024rethinking}, joint rotations~\cite{xiao2025motionstreamer,li2023object}) with an object state parameterized by global $SE(3)$ transformations (e.g., continuous 6D rotations~\cite{zhou2019continuity} plus 3D translations~\cite{lv2024himo,li2023object,li2023controllable,diller2023cghoi,ghosh2023imos,xu2023interdiff}).
This mixed representation introduces a geometric mismatch: human motion resides in Euclidean space while object pose lies on the non-Euclidean Lie group $SE(3)$, making it difficult for generative models to learn spatial correlations such as contact and coordinated motion.
Furthermore, $SE(3)$-based object parameterizations assume rigid-body motion and do not naturally extend to articulated objects.
Recent efforts have incorporated articulated objects by adding a 1D joint angle to the 6-DoF global pose~\cite{pi2025coda,zhang2025bimart}, but this requires explicit joint-type specification and does not extend to objects with multiple articulated parts or diverse joint mechanisms (e.g., prismatic, screw).
% ROG~\cite{xue2025guiding} samples surface keypoints from the object mesh to capture geometry, but uses only 24 skeletal joints on the human side and is limited to single-object interactions.
% ROG~\cite{xue2025guiding} selects boundary-focused and sparse keypoints from the object mesh (24 only) to provide a coarse representation of object geometry, but ignores hand motion and fine-grained contact on the object surface. 
% More importantly, it is limited to single-object interactions only.\hj{Double confirm the description of ROG is accurate.}
Uni-Inter~\cite{liu2025uniinter} addresses the representation mismatch by encoding humans, objects, and scenes into a shared voxel-based occupancy volume; however, it requires object trajectories as input and relies on discrete voxelization.
In contrast, we represent objects using surface keypoint trajectories that handle multi-object and articulated interactions without explicit joint-type specification, and adopt surface markers on SMPL-X~\cite{SMPL-X:2019} for the human body following~\cite{wu2022saga,zhang2025bimart}, placing both in Euclidean space.

\myparagraph{Human-Object Interaction Generation.}
Recent works incorporate scene geometry or object motion priors to guide human motion synthesis, improving the modeling of human-scene and human-object interactions~\cite{huang2023diffusion,zhao2023synthesizing,wang2022towards,wang2022humanise,zhang2024scenic,jiang2024autonomous,luo2024grasping}.
Meanwhile, there has been growing interest in text-driven HOI generation, where many approaches jointly synthesize human motion and dynamic object motion~\cite{diller2023cghoi,wang2023physhoi,li2023controllable,song2024hoianimator,xu2024interdreamer,ron2025hoidini,peng2023hoi,xu2025interact}.
However, most existing methods focus on interactions with a single object in isolation.
HIMO~\cite{lv2024himo} is among the first to explore text-driven multi-object HOI generation, but requires the number of interacting objects to be pre-specified, limiting flexibility.
For articulated objects, existing approaches either restrict dynamics to revolute articulation~\cite{pi2025coda,zhang2025bimart}, covering only a narrow subset of everyday mechanisms, or focus on bimanual hand manipulation without synthesizing full-body motion~\cite{cha2024text2hoi,zhang2025bimart,huang2025hoigpt}.
Our method handles a variable number of rigid and articulated objects within a single model, enabling whole-body HOI generation across diverse interaction settings.

\myparagraph{Articulated and Multi-Object Interactions Generation.}
HIMO~\cite{lv2024himo} is among the first to explore text-driven multi-object HOI generation; however, it requires the number of interacting objects to be pre-specified, which limits modeling flexibility. 
For complex articulated objects, existing approaches either restrict object dynamics to revolute (hinge) articulation~\cite{pi2025coda,zhang2025bimart}, covering only a narrow subset of everyday mechanisms, or focus on bimanual hand manipulation without synthesizing full-body motion~\cite{cha2024text2hoi,zhang2025bimart,huang2025hoigpt}. 
In contrast, our method enables whole-body HOI generation with diverse object dynamics in richer multi-object settings, yielding more realistic and flexible human--object interaction synthesis.

\begin{figure*}[!t]
    \centering
    \begin{overpic}[width=1.0\linewidth]{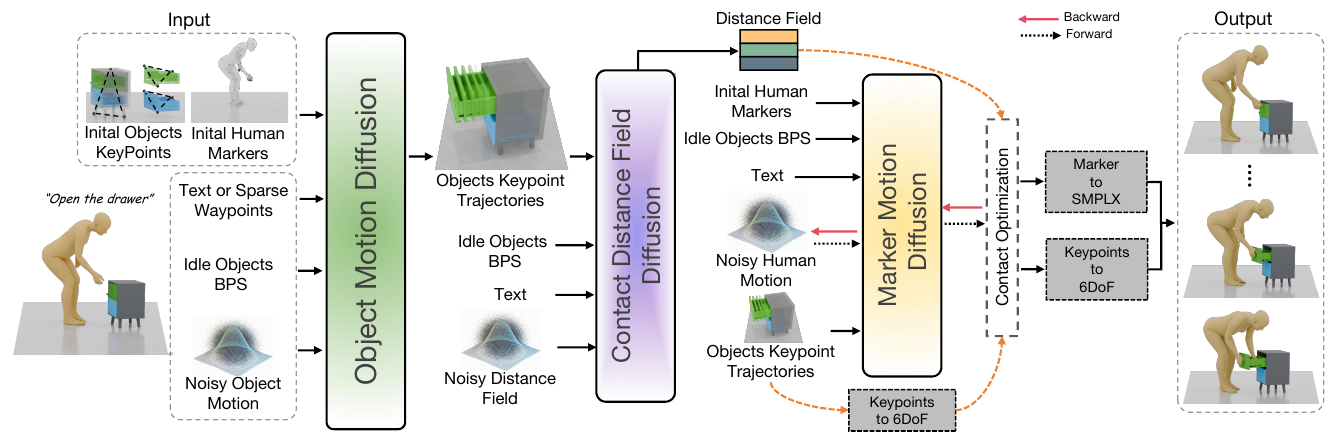}
     \put(144,8){\small{$f^{o}$}}
     \put(248,18){\small{$f^d$}}
      \put(350,28){\small{$f^h$}}
     \put(60,-10) {\small  \textbf{Stage1: Object Motion Generation} \quad  \textbf{Stage2: Contact Distance Field Prediction} \quad \textbf{Stage3: Body Motion Synthesis}}
 \end{overpic}
    % \vspace{0.01em}
    % \vspace{2pt}
    \caption{\textbf{Overview of the proposed framework.} 
    Our pipeline consists of three stages. 
    \textbf{Stage 1 (Object Motion Generation)} employs an object-motion diffusion model $f^o$ to generate object surface keypoint trajectories conditioned on the initial HOI state, the text prompt, and object geometry encoded with Basis Point Sets (BPS). 
    \textbf{Stage 2 (Contact Distance Field Prediction)} uses a diffusion model $f^c$ to predict a human--object contact distance field based on the generated object motion, the text prompt, and the object geometry.
    \textbf{Stage 3 (Body Motion Synthesis)} applies a marker-motion diffusion model $f^h$ to synthesize the final human motion conditioned on the text prompt, the generated object motion, and the predicted contact distance field. 
    We further perform contact optimization to alleviate contact and penetration artifacts, and finally recover the human and object meshes.
    % \hj{Update the visualization of the human body.}
    }
    \label{fig:framework}
    \vspace{-1.5em}
\end{figure*}
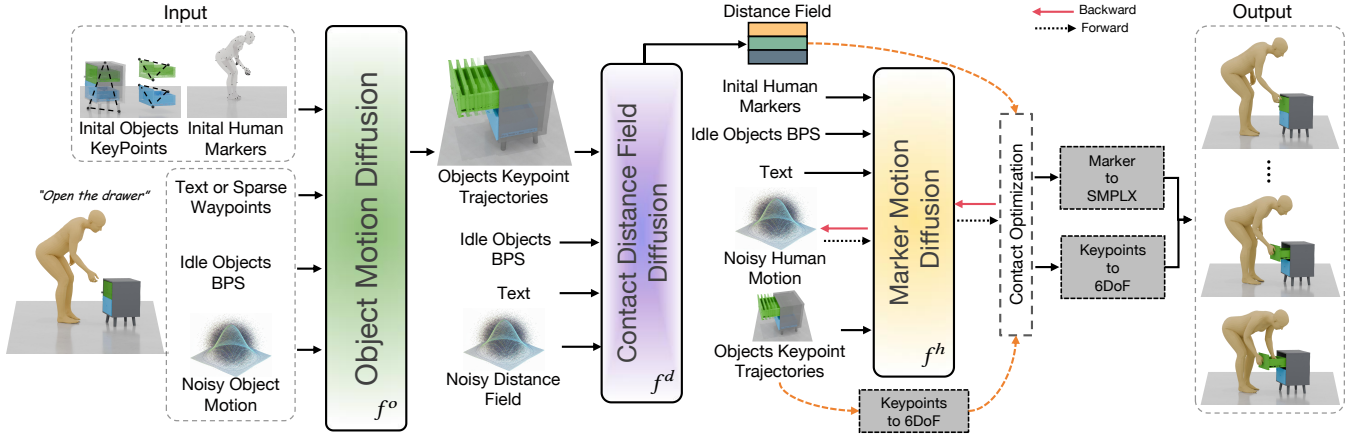

\myparagraph{Contact Modeling for HOI Generation.}
Modeling contact between the human body and objects is essential for physically plausible interaction synthesis.
Early approaches predict binary contact maps or contact likelihoods over the human body~\cite{petrov2024tridi,li2023controllable,diller2023cghoi}, object surfaces~\cite{cha2024text2hoi,jian2023affordpose,chu20253d,zhang2025bimart}, or both~\cite{tripathi2023deco,yang2024lemon}, but such binary signals lack the spatial precision to recover which body part touches which object region.
Several works explore richer contact representations: some model static paired contact patterns~\cite{liu2023contactgen,cseke2025pico}, while others capture dynamic correspondences only for a limited set of body parts (e.g., bimanual hand manipulation)~\cite{li2023controllable,ron2025hoidini,pi2025coda}.
ManipNet~\cite{he2021manipnet} introduces distance-based spatial sensors between hands and nearby object surfaces, but these are local to the hand and restricted to single-object settings.
ROG~\cite{xue2025guiding} constructs an Interactive Distance Field between 24 skeletal keypoints and object surface keypoints with spatial and temporal attention, representing the closest prior work to ours.
However, ROG handles only single-object interactions, and computing distances from interior skeletal joints rather than the body surface limits contact precision.
In contrast, our spatio-temporal contact distance field extends distance-based contact modeling to whole-body, multi-object, and articulated settings, computing distances from 138 body surface markers to 384 surface points per rigid component.

\vspace{-1mm}
\section{Methodology}

% An overview of our framework is shown in Fig.~\ref{fig:framework}.

\subsection{Surface Keypoint Representation}
\label{med:representation}
 
We represent object motion using surface keypoint trajectories and human motion using SMPL-X surface markers, placing both in Euclidean space so that spatial relationships such as contact and proximity reduce to point-to-point distance computations.
 
\myparagraph{Object Representation.}
Most existing methods represent object states using global $SE(3)$ transformations, which assume rigid-body motion and require explicit joint-type specification when extended to articulated objects~\cite{pi2025coda, zhang2025bimart}.
Instead, we model each scene as a collection of rigid components, where a rigid object constitutes a single component and an articulated object consists of multiple components (\eg, a drawer unit has a fixed body and several movable drawers, each as a separate component).
For each rigid component, we sample $K$ non-collinear surface keypoints from its canonical mesh using farthest point sampling~\cite{qi2017pointnet++}, where $K \geq 3$ is the minimum required to uniquely determine a rigid transformation.
Because each component is rigid, $K$ non-collinear points fully determine its pose, making the representation lossless for rigid motion while operating entirely in Euclidean space and avoiding the manifold constraints of $SE(3)$ parameterizations.
At each time step $t$, the object state is represented by the global 3D positions of these keypoints.
A key property of this representation is that different articulation mechanisms (revolute, prismatic, and screw joints) each produce distinctive keypoint trajectory patterns (as illustrated in Fig.~\ref{fig:articulated}), which a generative model can learn directly from data without requiring explicit joint-type specification.
 
In addition, existing multi-object methods require pre-specifying the number of objects at training time~\cite{lv2024himo}.
To handle a variable number of rigid components, we define a maximum capacity $N$ and represent the full object state at time $t$ as $\mathbf{O}_t \in \mathbb{R}^{N\times 3K}$.
When fewer than $N$ components are present, unused slots are padded with zeros and masked out during training and inference.
The rigid transformation of each component can be recovered in closed form: given the canonical keypoints $\mathbf{O}_0^i \in \mathbb{R}^{3K}$ and predicted keypoints $\mathbf{O}_t^i \in \mathbb{R}^{3K}$ for the $i$-th rigid component, we use the Kabsch algorithm~\cite{Kabsch:a12999} to recover its rotation and translation as
\begin{equation}
    \mathbf{R}_t^i, \mathbf{t}_t^i = \argmin_{\mathbf{R}, \mathbf{t}} \| \mathbf{R} \mathbf{O}_0^i + \mathbf{t} - \mathbf{O}_t^i \|^2.
\end{equation}
 
\myparagraph{Human Representation.}
Following prior work~\cite{wu2022saga, zhang2025bimart}, we represent the human body using $M$ surface markers sampled on the SMPL-X body mesh~\cite{SMPL-X:2019}, including dense markers on the palms for fine-grained hand contact.
At each time step $t$, the human state is represented by the global 3D positions of these markers, denoted as $\mathbf{H}_t \in \mathbb{R}^{M \times 3}$.
We use $M = 138$ markers in our experiments.
For rendering and evaluation, we train a lightweight fitting model that maps predicted marker positions to SMPL-X parameters, enabling recovery of the full human mesh. Please refer to the Appendix for more details.
 
We note that while both humans and objects are represented as surface points, the object keypoint representation exploits a property specific to rigid bodies: a small number of non-collinear points fully determine pose, enabling lossless, joint-type-agnostic motion modeling that has no analog in the deformable human setting.
These design choices are validated empirically in Sec.~\ref{sec:ablation}.

% \begin{figure*}
%     \centering
%     \begin{overpic}[width=1.0\linewidth]{figures/contact_vis_v3.pdf}
%      % \put(9,120){\small{w/o Contact}}
%      %  \put(9,110){\small{Optimization}}
%      % \put(9,210){\small{w/o Separate}}
%      %  \put(16,200){\small{Stage}}
%      % \put(9,30){\small{Ours (Full)}}
%     \end{overpic}
%     \vspace{-2em}
% \caption{\textbf{Contact visualization based on the predicted distance field.} 
% % We visualize contacts using the predicted contact field by extracting contact points via thresholding, and highlight the corresponding points on both the human and the object with red markers.
% From left to right, we show key frames of the generated results over time. We render rays between human markers and object surface points, both shown as white spheres, with different styles to indicate contact strength. No ray is rendered when the distance field value is below 0.8, indicating weak human--object proximity. Values between 0.8 and 0.9 are visualized as weak contact using semi-transparent colored rays, while values above 0.9 are visualized as strong contact using red rays.
% }
%     \label{fig:contact_vis}
%     \vspace{-1em}
% \end{figure*}

\begin{figure*}[t]
    \centering
    \setlength{\tabcolsep}{1pt}
    \begin{tabular}{cccc}
    \multicolumn{4}{c}{\textbf{Textual prompt:} Open the cabinet.} \\
         \includegraphics[width=0.24\linewidth]{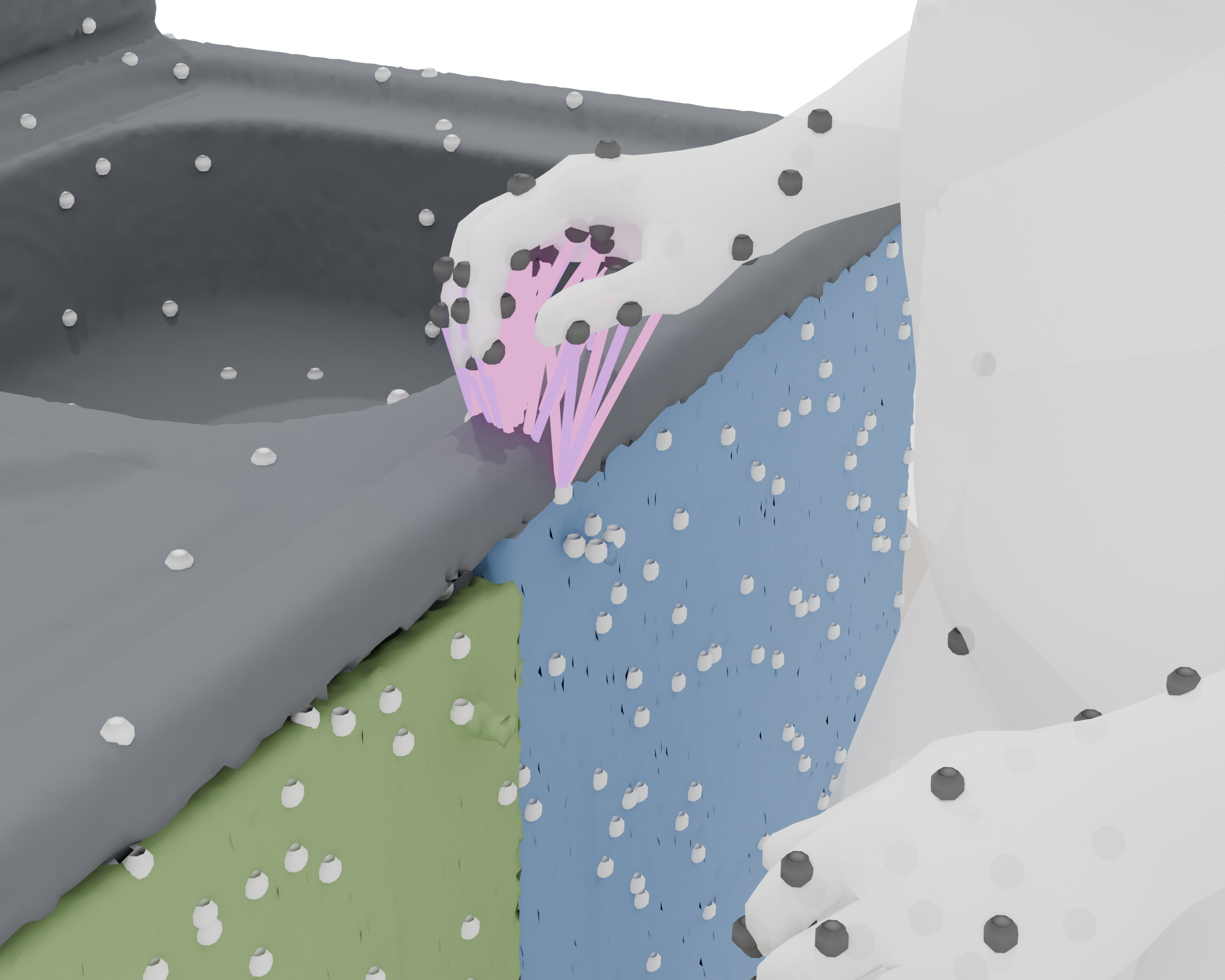} &
        \includegraphics[width=0.24\linewidth]{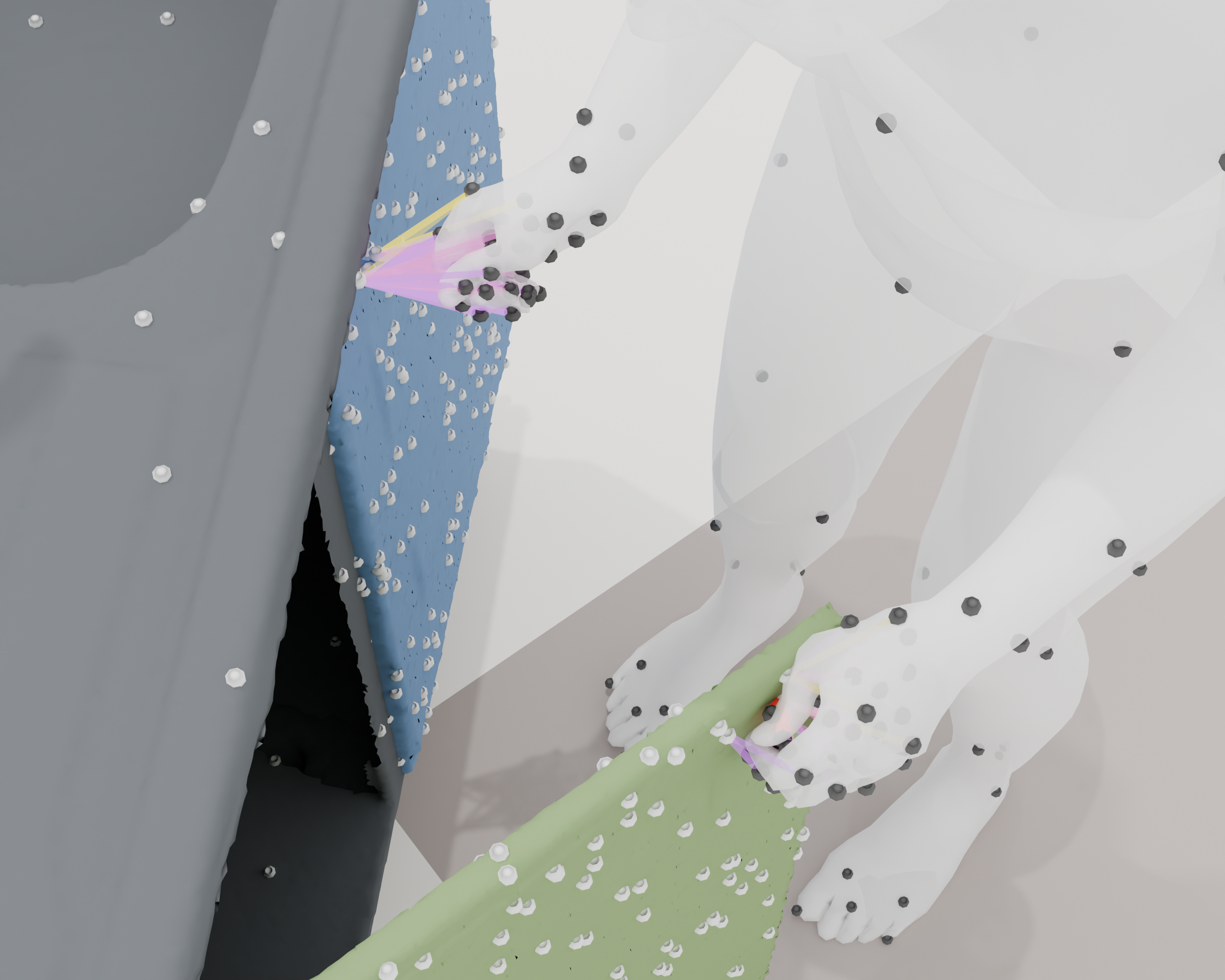} &
        \includegraphics[width=0.24\linewidth]{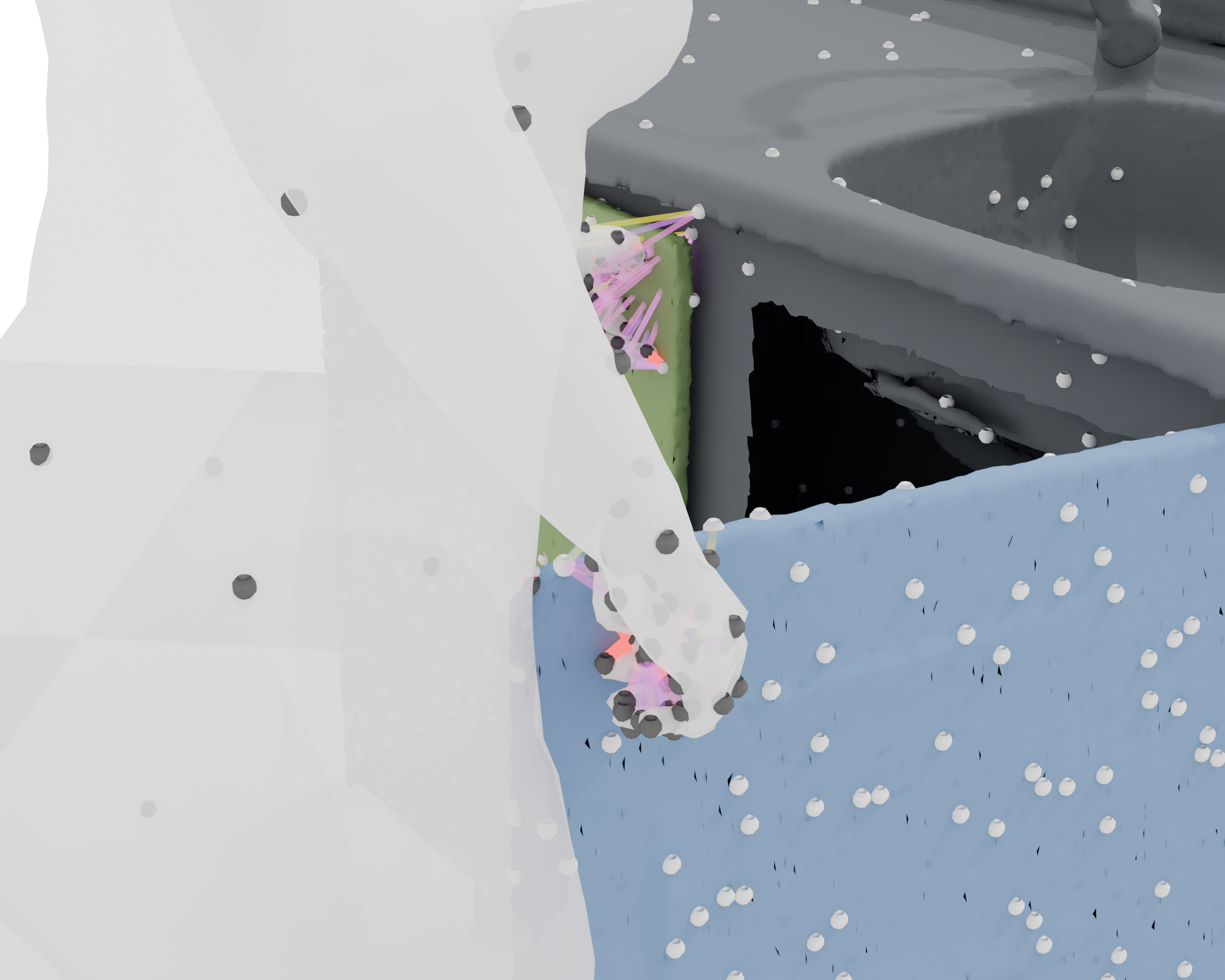} &
        \includegraphics[width=0.24\linewidth]{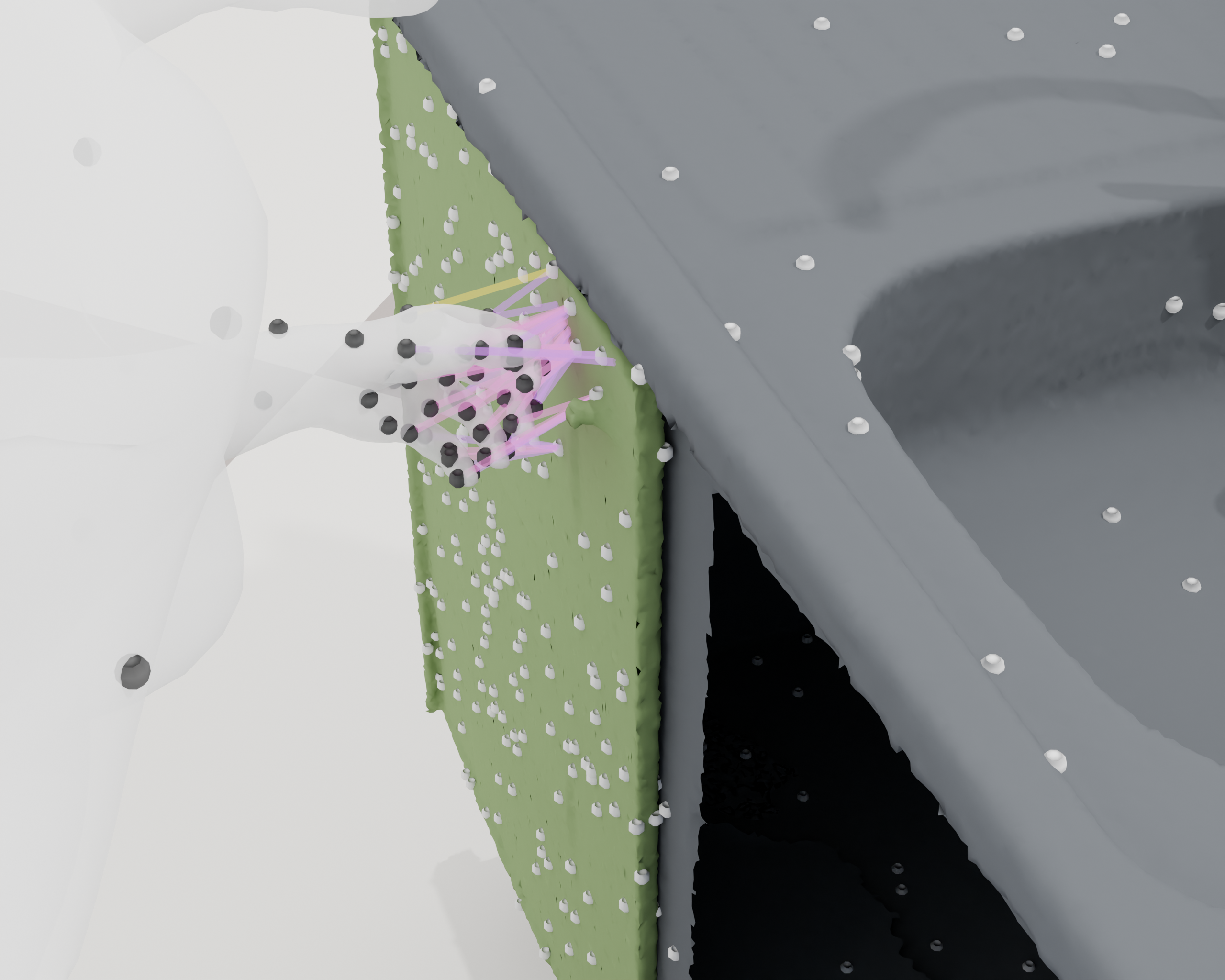} \\
    $t=4$ & $t=43$ & $t=83$ & $t=123$ \\
    \end{tabular}
    % \end{tabular}
    % \begin{subfigure}{0.24\linewidth}
    %     \centering
    %     \includegraphics[width=\linewidth,trim=20cm 20cm 20cm 10cm,
    % clip]{figures/contact_vis/frames_contact_270/frame_0000.png}
    %     % \caption{}
    % \end{subfigure}
    % \hfill
    % \begin{subfigure}{0.24\linewidth}
    %     \centering
    %     \includegraphics[width=\linewidth]{figures/contact_vis/frames_contact_270/frame_0040.png}
    %     % \caption{}
    % \end{subfigure}
    % \hfill
    % \begin{subfigure}{0.24\linewidth}
    %     \centering
    %     \includegraphics[width=\linewidth]{figures/contact_vis/frames_contact_270/frame_0080.png}
    %     % \caption{}
    % \end{subfigure}
    %     \hfill
    % \begin{subfigure}{0.24\linewidth}
    %     \centering
    %     \includegraphics[width=\linewidth]{figures/contact_vis/frames_contact_270/frame_0123.png}
    %     % \caption{}
    % \end{subfigure}
    %     \hfill
    % \begin{subfigure}{0.32\linewidth}
    %     \centering
    %     \includegraphics[width=\linewidth]{figures/600x400.png}
    %     % \caption{}
    % \end{subfigure}
    %     \hfill
    % \begin{subfigure}{0.32\linewidth}
    %     \centering
    %     \includegraphics[width=\linewidth]{figures/600x400.png}
    %     % \caption{}
    % \end{subfigure}
    \vspace{-12pt}
\caption{\textbf{Contact visualization based on the predicted contact distance field.} 
% We visualize contacts using the predicted contact field by extracting contact points via thresholding, and highlight the corresponding points on both the human and the object with red markers.
From left to right, we show key frames of the generated results over time. 
Human markers and object surface points are shown as white and blue dots, respectively.
To visualize the predicted contact distance field, we draw lines between marker-surface pairs whose field value exceeds 0.8: values between 0.8 and 0.9 are shown as semi-transparent colored lines (weak contact), while values above 0.9 are shown as solid red lines (strong contact). Pairs below 0.8 are not drawn.}
    \vspace{-12pt}
    \label{fig:contact_vis}
\end{figure*}

\subsection{Factorized HOI Generation}
\label{med:threestage}
Jointly modeling object motion, contact dynamics, and full-body kinematics is highly complex, making it difficult for a single end-to-end model to learn all components well.
We therefore decompose the problem into three sequential stages: object motion generation, contact distance field prediction, and human body motion synthesis.
Fig.~\ref{fig:framework} illustrates our pipeline.
% Specifically, \textbf{Stage~I} generates object keypoint trajectories given an instruction (either a text prompt or a set of waypoints) and the initial states of objects and human body.
% \textbf{Stage~II} then predicts a spatio-temporal contact distance field conditioned on the generated object motions.
% \textbf{Stage~III} finally synthesizes whole-body motion conditioned on the object trajectories, and refines human-object contacts via optimization guided by the predicted distance field.
Specifically, we formulate each stage as a conditional diffusion model based on the Transformer architecture~\cite{vaswani2017attention}.
% which has proven effective for generating diverse and realistic human motions~\cite{tevet2023human, chen2023executing}.
% All three stages share the same Transformer-based architecture~\cite{vaswani2017attention}, taking noisy input tokens and conditioning signals to output denoised predictions.
We adopt Flow Matching~\cite{lipman2022flow, rectifiedflow} for its efficacy and training simplicity, minimizing the mean squared error (MSE) between predicted and ground-truth velocities in a latent space.
For brevity, we will explain how the diffusion models work in the Euclidean space in the rest of this section, which can be easily extended to the latent space.
More details are provided in the supplementary material.

\noindent{\textbf{Stage I: Object Motion Generation.}}
Stage~I predicts object keypoint trajectories $\hat{\mathbf{O}}_{1:T}=\{\hat{\mathbf{O}}_t\}_{t=1}^T \in \mathbb{R}^{T\times N \times 3K}$ for all $N$ rigid components across $T$ time steps\footnote{Throughout this paper, we use $\hat{\cdot}$ (\eg, $\hat{\mathbf{O}}_{1:T}$) to denote quantities that are predicted by or depend on a neural network's output.}.
The generation is conditioned on four inputs: (1) an instruction embedding $\mathbf{e}$ derived from either a text prompt or sparse waypoints; (2) per-object Basis Point Set (BPS) features $\mathbf{b}$ that encode each object's geometry in its canonical pose~\cite{prokudin2019efficient}; (3) initial human markers $\mathbf{H}_{\mathrm{init}} \in \mathbb{R}^{L_{\mathrm{init}} \times 3M}$; and (4) initial object keypoints $\mathbf{O}_{\mathrm{init}} \in \mathbb{R}^{ L_{\mathrm{init}} \times N \times 3K}$, where $L_{\mathrm{init}}$ denotes the number of initial frames (4 in our implementation) that specify the HOI's starting location.

We denote the noisy object motion as $\boldsymbol{\epsilon}^o \in \mathbb{R}^{T\times N \times 3K}$.
We flatten the object and time dimensions into $TN$ tokens, enabling our Transformer diffusion model $f^o$ to handle a variable number of objects.
The object trajectories are generated as
\begin{equation}
    \hat{\mathbf{O}}_{1:T} = f^{o}(\mathbf{e},\, \mathbf{b},\, \mathbf{H}_{\mathrm{init}},\, \mathbf{O}_{\mathrm{init}},\, \boldsymbol{\epsilon}^o).
    \label{eq:stage1_gen}
\end{equation}
$\hat{\mathbf{O}}_{1:T}$ will serve as conditioning for both Stage~II and Stage~III. 

\noindent{\textbf{Stage II: Contact Distance Field Prediction.}}
Before generating human motion, Stage~II predicts a spatio-temporal contact distance field that captures when and where body markers should contact each object, as shown in \Fref{fig:contact_vis}.
By decoupling contact prediction from motion synthesis, we allow the model to reason explicitly about body-object correspondences, which is crucial for multi-object and multi-part interactions where different body parts may engage with different objects or parts at different times.

\begin{table*}[t]
\renewcommand{\arraystretch}{1.0}
\begin{center}
    \resizebox{1.0\linewidth}{!}{\begin{tabular}{l | l |ccccc | ccc| ccccc | ccc}
       \toprule
        & & \multicolumn{8}{c|}{2 objects/components} & \multicolumn{8}{c}{3 objects/components} \\
        \cmidrule{3-10}\cmidrule{11-18}
           Dataset & Method& \multicolumn{5}{c|}{Motion} & \multicolumn{3}{c|}{Interaction} &\multicolumn{5}{c|}{Motion} & \multicolumn{3}{c}{Interaction}\\
        \cmidrule{3-7}\cmidrule{8-10}\cmidrule{11-15}\cmidrule{16-18}
          & \ &  FID $\downarrow$  & $R_{prec}$ $\uparrow$ & Div $\rightarrow$ & FS $\downarrow$ & Jerk$_{obj}$ $\downarrow$ & $C_{acc}^{tem}\uparrow$ & $C_{acc}^{body}$$\uparrow$ & Pene $\downarrow$  &  FID $\downarrow$  & $R_{prec}$ $\uparrow$ & Div $\rightarrow$ & FS $\downarrow$ & Jerk$_{obj}$ $\downarrow$ & $C_{acc}^{tem}\uparrow$ & $C_{acc}^{body}$$\uparrow$ & Pene $\downarrow$ \\
        \midrule
        \multirow{3}{*}{\rotatebox[origin=c]{45}{ParaHome}}
          & Real motion (reference)             &  0.00  & 0.727  & 7.78   & 0.0039 & 0.15 & - &  -    & -                            & 0.00    &  0.679  & 7.47  & 0.0031  & 0.05     & -     &  -   &- \\
          & HIMO-Gen~\cite{lv2024himo}            &  14.83  & 0.580  & 8.23   & 0.3258 & 8.15 & 0.534 & 0.815 & 0.570    & 21.61   &  0.509  & 7.03  &  0.4437 & 7.13 & 0.634 & 0.809 & 0.857 \\
          & Ours   & \textbf{4.49}    &\textbf{0.707}  & 8.38 & \textbf{0.0035} & \textbf{0.72} &\textbf{0.669}& \textbf{0.896}  &  \textbf{0.536}             & \textbf{6.09} & \textbf{0.598} & \textbf{7.57} & \textbf{0.0017}     & \textbf{0.40}   & \textbf{0.680}   &  \textbf{0.906}  & \textbf{0.776}\\
        \midrule
        \multirow{3}{*}{\rotatebox[origin=c]{45}{HIMO}}
          & Real motion  (reference)            &  0.000  & 0.729  & 11.905   & 0.0007 & 0.08 & - &  -    & -                            & 0.267    &  0.713  & 9.755  & 0.0004  & 0.10     & -     &  -   &- \\
          & HIMO-Gen~\cite{lv2024himo}            &  8.019  & 0.570  & 10.387   & 0.0021 & 0.89 & 0.612 & 0.698 & 0.670    & 4.467   &  0.556  & \textbf{9.777}  &  0.0038 & 0.43 & 0.656 & 0.712 & \textbf{0.563}\\
          & Ours   & \textbf{6.572}    &\textbf{0.612}  & \textbf{11.021} & \textbf{0.0011} & \textbf{0.17} & \textbf{0.731} & \textbf{0.912}  &  \textbf{0.519}       & \textbf{2.273} & \textbf{0.628} & 9.812 &\textbf{0.0015}     &\textbf{0.19}   & \textbf{0.751}   &  \textbf{0.883}  & 0.621\\
        \bottomrule
\end{tabular}}
\caption{\textbf{Quantitative results on ParaHome and HIMO for multi-object and articulated HOI generation.} We compare against HIMO-Gen~\cite{lv2024himo}. Best results are highlighted in \textbf{bold}. Note that lower penetration (Pene) does not always imply better contact, as it may also indicate no contact at all.
% \hj{Shall we say 2 objects/components as there are articulated objects in ParaHome? And ``motion'' is about ``human motion''?} \xg{2 object/components make sense for me, but there is a object metric (Jerk$_{obj}$) in motion section}
}
\label{table:quant1}
\end{center}
\vspace{-2.5em}
\end{table*}

We define the contact distance field between a subset of human markers and dense object surface points.
For the human, we select $M_c$ markers from the full set of $M$ markers, covering body regions that frequently participate in interactions (\eg, hands, torso), and denote them as $\mathbf{H}^c_t \in \mathbb{R}^{M_c \times 3}$ at time $t$.
For objects, we use $Q$ denser surface points per rigid component ($Q \gg K$) to capture finer geometric detail, and denote them as $\mathbf{S}_t \in \mathbb{R}^{NQ \times 3}$ at time $t$.
The contact distance field $\mathbf{D}_t \in \mathbb{R}^{M_c \times NQ}$ encodes the proximity between each marker-surface pair (1 indicates contact and 0 not), computed as
\begin{equation}
    \mathbf{D}_t = \sigma\!\left(\frac{\tau - \mathtt{dist}(\mathbf{H}^c_t, \mathbf{S}_t)}{\alpha}\right),
    \label{eq:contact_gt}
\end{equation}
where $\mathtt{dist}(\cdot, \cdot)$ computes pairwise Euclidean distances, $\sigma(\cdot)$ is the sigmoid function, $\tau$ is a contact threshold, and $\alpha$ controls sharpness.
 
This formulation differs from prior contact representations in several important ways.
Binary contact labels~\cite{diller2023cghoi, peng2023hoi} are extremely sparse, resulting in weak gradients that make it difficult for the model to learn fine-grained contact transitions.
ManipNet~\cite{he2021manipnet} introduces distance-based spatial sensors between hands and objects, but these are local to the hand and restricted to single-object settings.
ROG~\cite{xue2025guiding} proposes an Interactive Distance Field between skeleton joints and object boundary keypoints, but does not model fine-grained contact and is also limited to a single object.
Signed distance fields (SDFs) encode inside/outside geometry, which is useful for penetration penalties but does not directly capture interaction-aware proximity between body and object surfaces over time.
In contrast, our contact distance field is fully spatio-temporal, fine-grained, and captures whole-body correspondences across multiple objects.
We validate this choice empirically in Sec.~\ref{sec:ablation}.
Visual results are shown in Fig.~\ref{fig:contact_vis}.
 
Given initial noise $\boldsymbol{\epsilon}^d \in \mathbb{R}^{T \times M_c \times NQ}$ and the generated object trajectories $\hat{\mathbf{O}}_{1:T}$ from Stage~I, our Contact Distance Field Diffusion model predicts
\begin{equation}
    \hat{\mathbf{D}}_{1:T} = f^d(\mathbf{e},\, \mathbf{b},\, \hat{\mathbf{O}}_{1:T},\, \boldsymbol{\epsilon}^d).
    \label{eq:stage2_gen}
\end{equation}
Similar to Stage~I, we flatten the marker and time dimensions into $TM_c$ tokens for the Transformer.
The predicted $\hat{\mathbf{D}}_{1:T}$ serves as contact priors for optimization in Stage~III.

\noindent{\textbf{Stage III: Human Motion Synthesis with Contact Optimization.}}
The goal here is to generate the human motion sequence $\hat{\mathbf{H}}_{1:T} = \{\hat{\mathbf{H}}_t\}_{t=1}^T \in \mathbb{R}^{T \times M \times 3}$ conditioned on the predicted object trajectories $\hat{\mathbf{O}}_{1:T}$ from Stage~I, and further refine the motion using the predicted contact distance field $\hat{\mathbf{D}}_{1:T}$ from Stage~II.
Given initial noise $\boldsymbol{\epsilon}^h \in \mathbb{R}^{T \times M\times 3}$, our Body Motion Diffusion model generates
\begin{equation}
    \hat{\mathbf{H}}_{1:T} = f^h(\mathbf{e},\, \mathbf{b},\, \mathbf{H}_{\mathrm{init}},\, \mathbf{O}_{\mathrm{init}},\, \hat{\mathbf{O}}_{1:T},\, \boldsymbol{\epsilon}^h).
    \label{eq:stage3_gen}
\end{equation}

While the diffusion model produces plausible motions, it may not precisely satisfy the predicted contacts, and directly optimizing $\mathbf{H}_{1:T}$ to minimize contact violations can push the result off the learned manifold, leading to unrealistic artifacts.
To address this, we adopt Diffusion Noise Optimization (DNO)~\cite{karunratanakul2024optimizing}, a technique also employed in recent HOI methods~\cite{pi2025coda, ron2025hoidini}, which optimizes the initial noise $\boldsymbol{\epsilon}^h$ rather than the output motion.
The key idea is to treat the diffusion model as a differentiable decoder: we define a loss measuring contact violations on the generated motion, backpropagate gradients through the entire denoising process, and iteratively update the noise to find a latent sample that decodes into motion with smaller contact errors.
Because the output is always decoded through the diffusion model, it remains on the learned motion manifold, preserving realism while improving contact accuracy.

We optimize the noise $\boldsymbol{\epsilon}^h$ by minimizing
\begin{equation}
    \mathcal{L} = \mathcal{L}_{\text{contact}} + \lambda_{\text{pen}} \mathcal{L}_{\text{pen}}
    \label{eq:total_loss}
\end{equation}
where $\mathcal{L}_{\text{contact}}$ encourages predicted contact pairs (identified by thresholding $\hat{\mathbf{D}}_t$) to be spatially close, and $\mathcal{L}_{\text{pen}}$ penalizes human markers that penetrate object interiors using the precomputed object SDF.
Visual examples of the predicted contacts are shown in Fig.~\ref{fig:contact_vis}.
The detailed loss formulations are provided in the Appendix.
At each denoising step, we decode the current noise to obtain $\hat{\mathbf{H}}_{1:T}$, compute the loss, and backpropagate through the diffusion model to update $\boldsymbol{\epsilon}^h$.
This optimization is applied at inference time to ensure contact accuracy.
When faster inference is desired, the optimization can be reduced or skipped entirely, with a modest trade-off in contact quality.
\section{Experiments}
\label{sec:experiment}

\subsection{Setup}
\label{sec:dataset}
\textbf{Datasets.} We use the ParaHome~\cite{kim2025parahome} dataset to evaluate HOI generation with multiple or articulated objects, which contains 486 minutes of motion sequences spanning 22 object categories.
We additionally evaluate on the HIMO~\cite{lv2024himo} benchmark for multi-object interactions and the ARCTIC benchmark~\cite{fan2023arctic} for articulated-object interactions, following their respective task settings.
We also train and evaluate our model on the single-object interaction dataset OMOMO~\cite{li2023object}, which includes human-object interaction motion for 15 objects, with a total duration of approximately 10 hours. We recover the hand motions for OMOMO following the method of~\cite{xu2025interact}.

\noindent\textbf{Evaluation metrics.} We evaluate different methods using the following metrics, considering both motion and interaction quality.

\begin{table}[t]
\renewcommand{\arraystretch}{1.1}
\begin{center}
    \resizebox{1.0\linewidth}{!}{\begin{tabular}{l |ccccccc}
       \toprule
       Method  &  FID $\downarrow$  & $R_{prec}$ $\uparrow$ & FS $\downarrow$ & Diversity  $\rightarrow$  & IV $\downarrow$  & ID $\downarrow$   & CR $\uparrow$  \\
               \midrule
         % \multicolumn{8}{c}{\cellcolor{SceneGreen}\text{ARCTIC Dataset}}  \\
         Real motion (reference)              & -  &  0.516   & 0.002   & 8.052 & 4.68 & 11.47 & 0.085 \\ \midrule
        CoDA~\cite{pi2025coda} & $\textbf{2.178}^{\pm.013}$  & $0.479^{\pm.003}$   &\textbf{0.003} & 7.562 & \textbf{5.25} & 12.87 & 0.086   \\
        HIMO-Gen~\cite{lv2024himo} & $3.064^{\pm.017}$  & $0.388^{\pm.002}$    &0.004 & 7.432 & 8.84 & 13.97 & 0.084   \\
         Ours   & $2.331^{\pm.014}$ & $\textbf{0.495}^{\pm.003}$  & \textbf{0.003}  & \textbf{7.892} & 5.37 & \textbf{11.98} & \textbf{0.089}  \\ 
         \bottomrule
\end{tabular}}
\vspace{-0.4mm}
\caption{\textbf{Quantitative results on the ARCTIC benchmark dataset for articulated-object HOI generation.} We compare our full method with our method's variants and the baseline CoDA~\cite{lv2024himo}. The best results are highlighted in \textbf{bold}.}
%\hzy{I noticed that we have a bad penetration score. Should we explain something like "if no contact occurs at all, then the penetration score would naturally to be low."?}
\label{table:quant_arctic}
\vspace{-2em}
\end{center}
\end{table}

\textit{Motion Quality.} We use \textbf{Fréchet Inception Distance}(FID) to quantify the distributional discrepancy between real and generated motions using a pretrained motion encoder. \textbf{R-Precision} (R$_{prec}$) measures the semantic alignment between generated motions and their corresponding text prompts. \textbf{Diversity} (Div) evaluates the range of variation across generated samples, reflecting motion richness. Additionally, we report the \textbf{Foot Sliding Score} (FS), computed as a weighted average of accumulated foot translation in the XY plane following prior work~\citep{karunratanakul2023guided}. The score is measured in centimeters (cm), with lower values indicating better generation quality. Finally, we report \textbf{Jerk$_{obj}$} to assess object motion quality, where lower jerk values indicate smoother and more natural movements.

\textit{Interaction Quality.} We assess the quality of human-object interactions by evaluating \textbf{Temporal Contact Accuracy} ($C_{acc}^{tem}$) and \textbf{Body Contact Accuracy} ($C_{acc}^{body}$), computed against the ground-truth spatial-temporal contact labels. To evaluate physical plausibility, we compute the \textbf{Penetration Score} (Pene) using the signed distance field (SDF) of the object mesh, following the prior works~\citep{li2023controllable}. More details are provided in the supplementary material.

\subsection{Comparisons with Existing Methods}

\noindent\textbf{Baselines.}
We compare against HOI-Diff~\citep{peng2023hoi}, CHOIS~\citep{li2023controllable}, ROG~\cite{xue2025guiding}, and HOIDiNi~\cite{ron2025hoidini} for single-object interactions.
For multi-object scenarios, we follow the official HIMO-Gen~\citep{lv2024himo} implementation and train separate models for interactions with two and three objects, respectively, since it does not support a variable number of objects.
For fair comparisons, we train all baseline methods using their original representations with the same length of initial HOI state and a generation length of $T{=}120$. 

\noindent\textbf{Quantitative Results.}
\Tref{table:quant1} reports quantitative results on ParaHome and HIMO for multi-object and articulated interactions.
On ParaHome, our full model achieves the best overall performance, obtaining state-of-the-art FID and R-Precision, indicating high motion quality and strong text alignment.
Moreover, improved $C_{acc}^{tem}$, $C_{acc}^{body}$, and competitive Pene score demonstrate more physically plausible and coherent human-object interactions.
On HIMO, our single model outperforms HIMO-Gen, which requires separate models for 2-object and 3-object scenarios.
On ARCTIC (\Tref{table:quant_arctic}), we achieve competitive performance with CoDA~\cite{pi2025coda}, demonstrating that surface keypoint trajectories can handle articulated objects without explicit joint-type specification.

% NOTE: Verify that table:quant_arctic matches the label in tables/table_ACRTIC.tex. Update if different.

\begin{table}[t]
\renewcommand{\arraystretch}{1.1}
\begin{center}
    \resizebox{1.0\linewidth}{!}{\begin{tabular}{l |ccccc|ccc}
       \toprule
        % & \multicolumn{8}{c}{\cellcolor{SceneBlue}\text{OMOMO Dataset}}  \\

            \multirow{2}{*}{Method} & \multicolumn{5}{c|}{Motion} & \multicolumn{3}{c}{Interaction} \\
            \cmidrule{2-6}\cmidrule{7-9}
          \ &  FID $\downarrow$  & $R_{prec}$ $\uparrow$ & Div $\rightarrow$ & FS $\downarrow$  & Jerk$_{obj}$ $\downarrow$ & $C_{acc}^{tem}\uparrow$ & $C_{acc}^{body}$$\uparrow$ & Pene $\downarrow$\\
        \midrule
         % & \multicolumn{8}{c}{\cellcolor{LightGreen}\textit{Single Rigid Object}}  \\
         Real motion (reference)             &  0.00  & 0.566  & 8.28   & 0.0002 & 0.44  &  -     & -   & -     \\ \midrule
        HOI-Diff~\cite{peng2023hoi} & 11.94   & 0.318   & 6.13 & 0.2074  &  58.01 & 0.446 & \textbf{0.915} & \textbf{0.477} \\
         CHOIS~\cite{li2023controllable}   & 9.11   & 0.461  & 7.39 & 0.0047   & 73.07                & 0.589 & 0.905  & 0.543 \\

          ROG~\cite{xue2025guiding}   & \textbf{3.27}   & 0.483  & 7.87 & 0.0036   & 16.05                & 0.719 & 0.898  & 0.586 \\
        HOIDiNi~\cite{ron2025hoidini}   & 4.87  & 0.512  & 7.95 & 0.0047  & 7.95                & 0.698 & 0.828  & 0.671 \\
         \midrule
       % w$/$o Separate Stage           &  5.09   & 0.504   & 7.89  &\textbf{0.0002}  & 3.13  & 0.634 & 0.878   & 0.542      \\
       %   w$/$o Noise Optimization            & \underline{3.85}    & \underline{0.518}   & \textbf{8.17} & \textbf{0.0002} &  \textbf{2.91} & \underline{0.760} & 0.901  & 0.591    \\
           \textbf{Ours}    &3.46  & \textbf{0.529} &\textbf{7.98} & \textbf{0.0004} & \textbf{2.95} & \textbf{0.886}  & 0.896 & 0.613\\
        \bottomrule
\end{tabular}}
\vspace{-0.4mm}
\caption{\textbf{Quantitative results on the OMOMO dataset for single-object HOI generation.} We compare our full method with our method's variants and existing baselines. The best results are highlighted in \textbf{bold}. }
% \hzy{I noticed that we have a bad penetration score. Should we explain something like "if no contact occurs at all, then the penetration score would naturally to be low."?}
\label{table:omomo}
\end{center}
\vspace{-2em}
\end{table}

We further evaluate on the single-object OMOMO benchmark, where our approach shows better or comparable performance with other models, as shown in \Tref{table:omomo}.

\noindent\textbf{Qualitative Results.}
We present qualitative comparisons with baseline methods in \Fref{fig:vis_results}. The baselines often produce inaccurate contacts and unstable object motions, especially for complex articulated objects.

\noindent\textbf{Perceptual User Study.}
We conduct a perceptual user study comparing our method against baselines on both ParaHome and OMOMO.
For each sample, participants evaluate the generated animations on two aspects: \emph{text alignment} (how well the motion matches the text description) and \emph{interaction quality} (how natural and plausible the human-object interaction appears). Details are in the supplementary material.
As shown in \Fref{fig:user_study}, our method is consistently preferred on both criteria across both datasets.

% \begin{table}[t]
% \renewcommand{\arraystretch}{1.2}
% \begin{center}
%     \resizebox{1.0\linewidth}{!}{\begin{tabular}{l|l|ccc}
%        \toprule
%         % & \multicolumn{7}{c|}{\cellcolor{SceneOrange}\textit{OMOMO}}  \\

%             % Rep & \multicolumn{4}{a}{Motion} & \multicolumn{3}{c|}{Interaction} \\
%         Cond. & Object Motion Representation \ & Jerk$_{obj}$ $\downarrow$ & $T_{obj}$ $\downarrow$ & $O_{obj}$ $\downarrow$ \\
%         \midrule
%          % ===== Joint block =====
%         \multirow{3}{*}{Text} 
%         & 6D rot + trans & 4.51 & - &- \\
%         & 9D rot matrix + trans & 7.23 & - & - \\
%         & Keypoints        & \textbf{3.75} & - & - \\
        
%         \midrule  
        
%         % ===== Obj Only block (shaded) =====
        
%         \multirow{3}{*}{Waypoints}
%         % \rowcolor{gray!12}
%         &  6D rot + trans & 4.02 & 10.13 & 1.11 \\    
%         &  9D rot matrix + trans      & 4.58 & 9.78 & 1.05 \\
%         & Keypoints                  & \textbf{2.86} & \textbf{7.51} &\textbf{ 0.97} \\

%         \bottomrule
% \end{tabular}}
% % \vspace{-4mm}
% \caption{\textbf{Ablative results of different object motion representation on the OMOMO dataset.}}
% \vspace{-2em}
% \label{table:diff_obj_rep}
% \end{center}
% \end{table}

\begin{table}[t]
\renewcommand{\arraystretch}{1.1}
\begin{center}
    \resizebox{0.95\linewidth}{!}{\begin{tabular}{l|c|ccc}
       \toprule
        \multirow{2}{*}{Object Motion Representation} & Text & \multicolumn{3}{c}{Waypoints} \\
        \cmidrule{2-2}\cmidrule{3-5}
          & Jerk$_{obj}$ $\downarrow$ & Jerk$_{obj}$ $\downarrow$ & $T_{obj}$ $\downarrow$ & $O_{obj}$ $\downarrow$ \\
        \midrule
        6D rot + trans~\cite{peng2023hoi}        & 4.51 & 4.02 & 10.13 & 1.11 \\
        9D rot matrix + trans~\cite{li2023controllable} & 7.23 & 4.58 & 9.78 & 1.05 \\
        Keypoints (Ours)             & \textbf{3.75} & \textbf{2.86} & \textbf{7.51} & \textbf{0.97} \\
        \bottomrule
\end{tabular}}
\caption{\textbf{Ablative results of different object motion representation on the OMOMO dataset.} Best results are highlighted in \textbf{bold}.
}
\vspace{-2em}
\label{table:diff_obj_rep}
\end{center}
\end{table}

\subsection{Ablation Studies}
\label{sec:ablation}
We ablate the two representation choices central to our contributions: the surface keypoint representation for objects and the contact distance field formulation.
We also provide ablation studies about our factorized pipeline in the supplementary material.

\begin{figure*}[t]
\vspace{-1em}
\centering
\begin{minipage}[t]{0.5\linewidth}
    \centering
    \begin{overpic}[width=\linewidth]{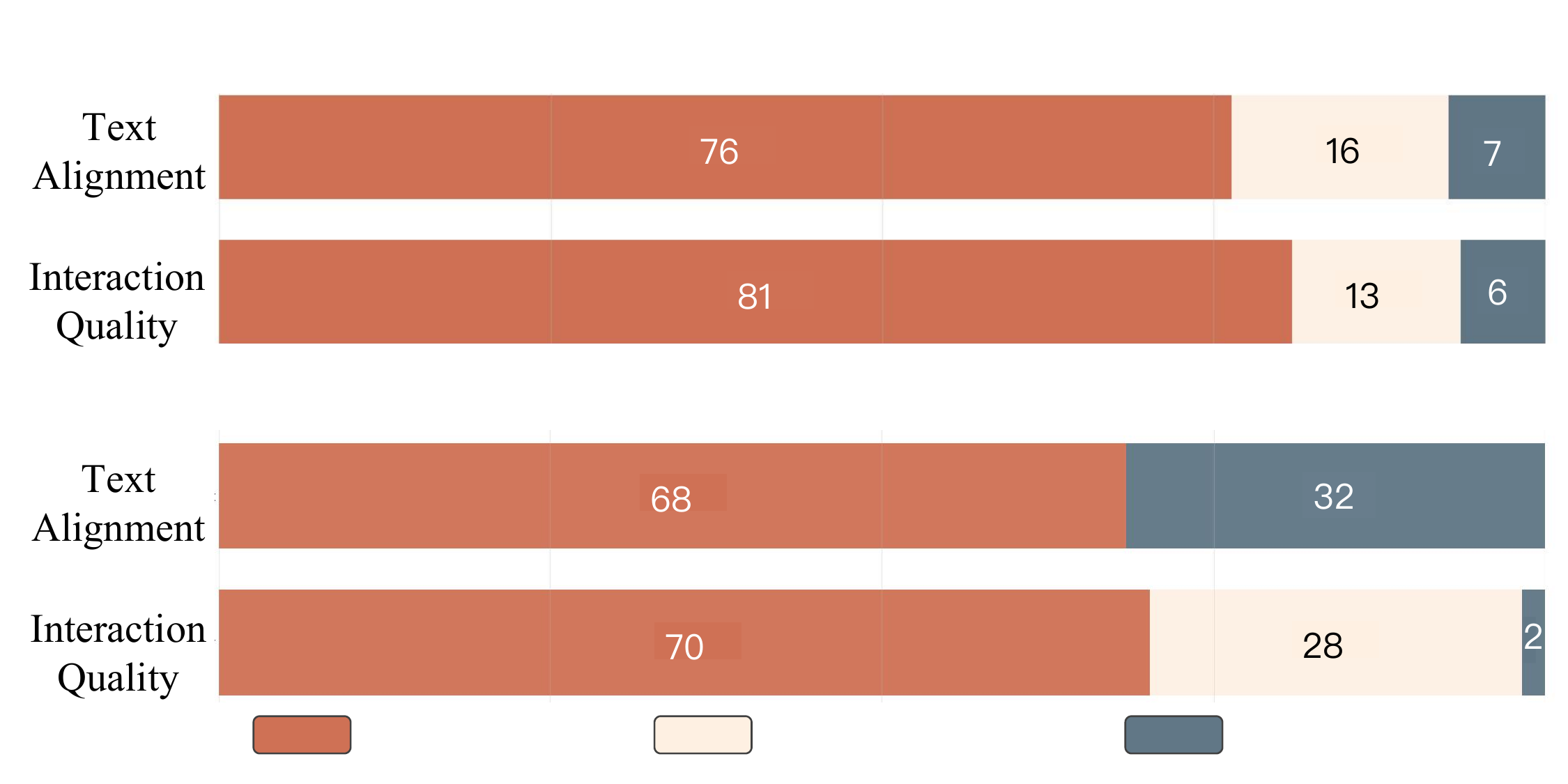}
      \put(58,5){\small{Prefer Ours}}
      \put(125,5){\small{Prefer Others}}
      \put(200,5){\small{Cannot tell}}
      \put(50,118){Interaction with Multiple or Articulated Objects}
      \put(70,60){Interaction with a Single Rigid Object}
    \end{overpic}
    \vspace{-2em}
    \caption{\textbf{User perceptual study results.} The percentage of times our approach is preferred over (a) Text Alignment, (b) Interaction Quality.}
    \label{fig:user_study}
\end{minipage}
\hfill
\begin{minipage}[t]{0.42\linewidth}
    \centering
    \begin{overpic}[width=\linewidth]{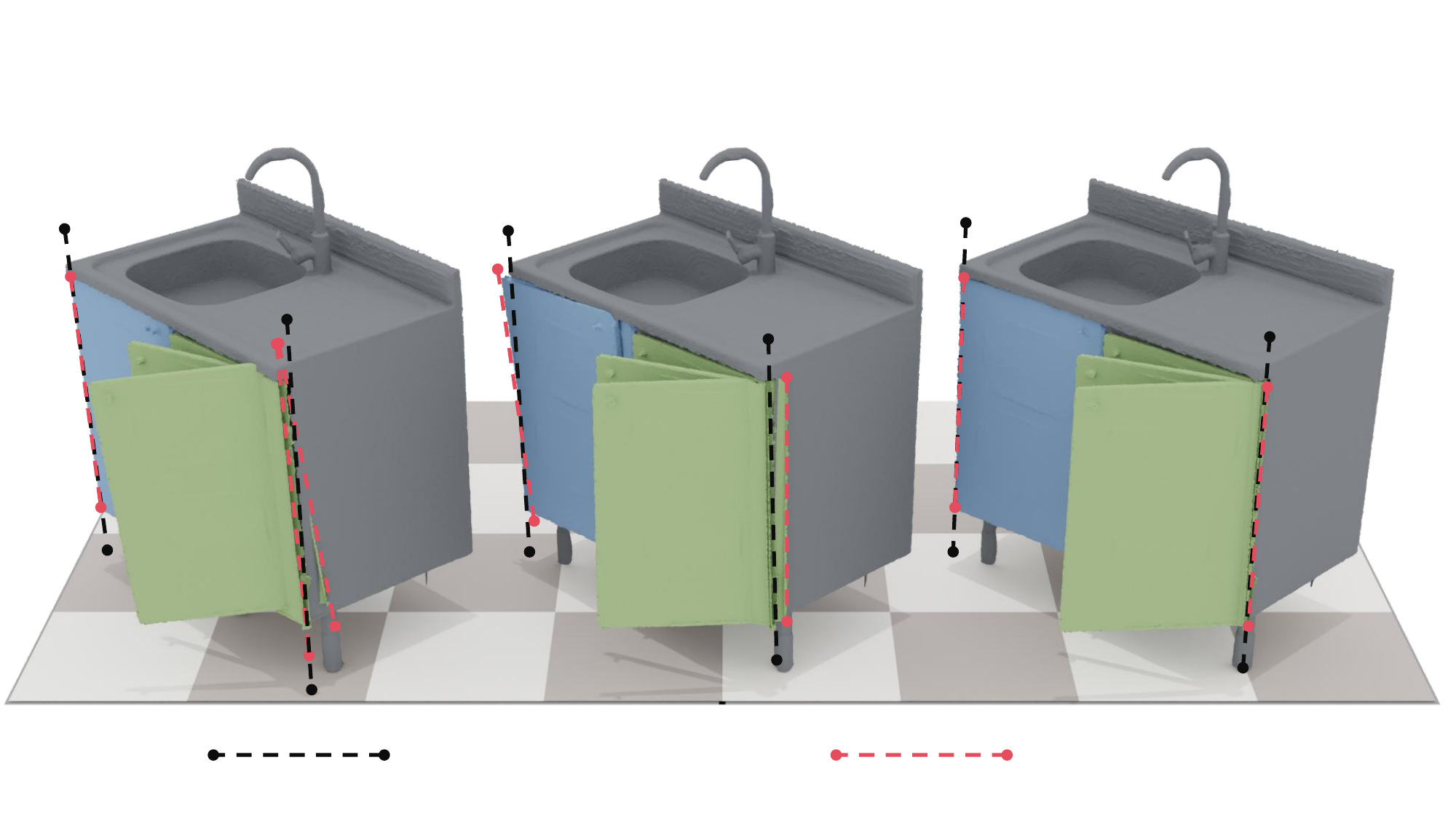}
      \put(2,110){\small{9D rot matrix + trans}}
      % \put(27,121){\small{+ trans}}
      \put(82,110){\small{6D rot + trans}}
      \put(155,110){\small{Keypoints}}
      \put(62,8){\small{Reference Axis}}
      \put(155,8){\small{Generated Axis}}
    \end{overpic}
    \vspace{-2em}
    \caption{\textbf{Point-based Representation \textit{vs.} Object Rotation + Translation.} Our \emph{surface keypoint representation} produces more realistic and accurate object motions, leading to more plausible object part motion in articulated scenarios.}
    \label{fig:rotation_3verts}
\end{minipage}
% \vspace{-1.5em}
\end{figure*}

\begin{figure*}[t]
% \vspace{-2em}
    \centering
     \begin{overpic}[trim={0 2 0 0}, width=0.8\linewidth]{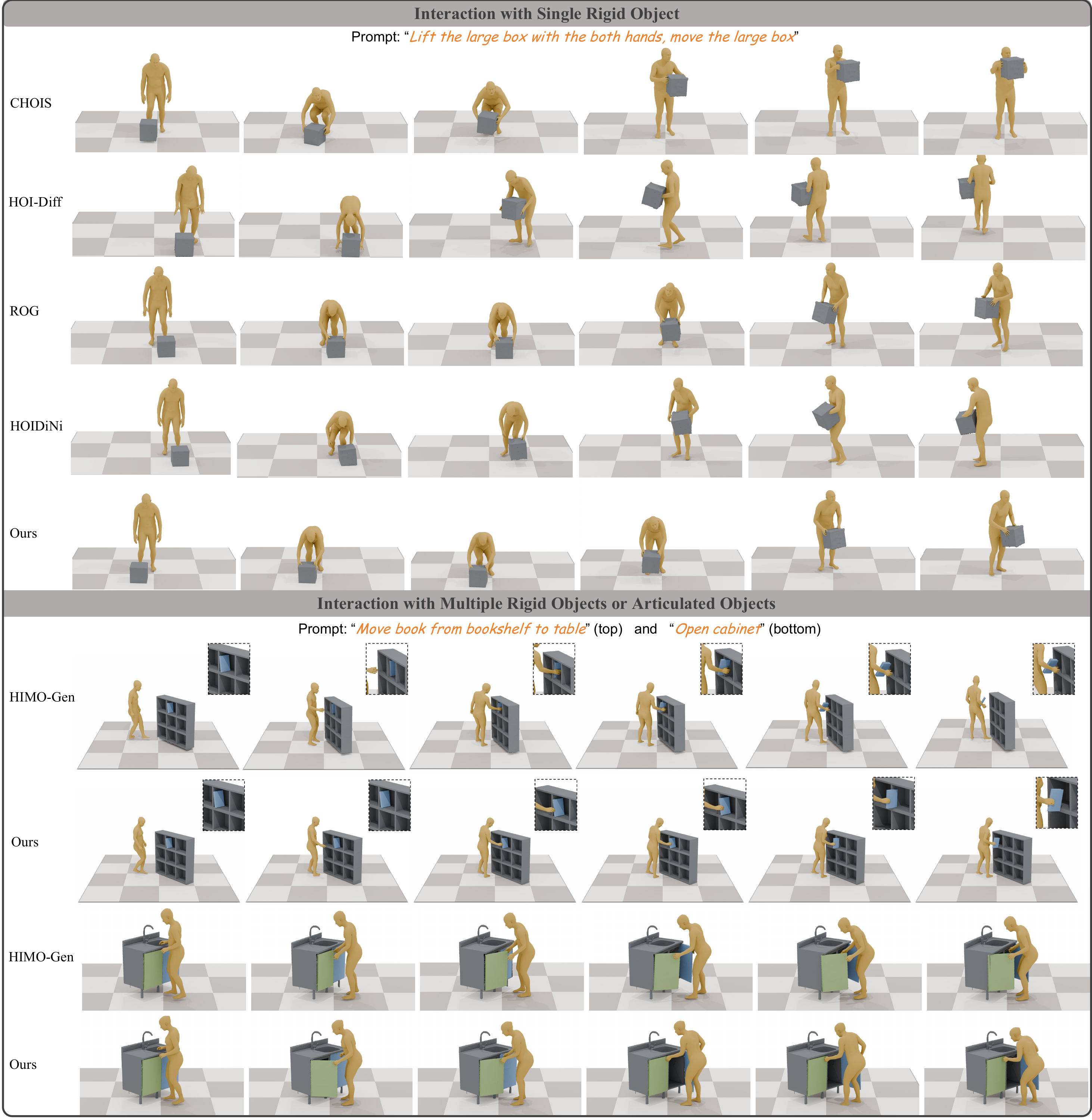}
    %  \put(3,470){\small{CHOIS}}
    %   \put(3,420){\small{HOI-Diff}}
    % \put(3,370){\small{ROG}}
    %   \put(3,320){\small{HOIDiNI}}
    %   \put(3,270){\small{Ours}}
    %  \put(3,180){\small{HIMO-Gen}}
    %  \put(3,130){\small{Ours}}
    %  \put(3,75){\small{HIMO-Gen}}
    %  \put(3,25){\small{Ours}}
    %  \put(188,513){\textbf{Interaction with a Single Rigid Object}}
    %   \put(165,237){\textbf{Interaction with Multiple or Articulated Objects}}
 
     \end{overpic}
    \vspace{-1.3em}
    \caption{\textbf{Qualitative comparisons on the test sets of two datasets (OMOMO: top; ParaHome: bottom).} For clarity, we visualize only keyframes from top to bottom for each interaction sequence. Compared to baseline methods, our approach generates more realistic and coherent human–object interactions, featuring more accurate contact and fewer artifacts.}
    \label{fig:vis_results}
    \vspace{-1em}
\end{figure*}

\myparagraph{Object Motion Representation.}
\Tref{table:diff_obj_rep} compares different object motion representations on OMOMO under text-only and waypoint conditioning~\cite{peng2023hoi,li2023object}, while keeping the Stage I architecture and training settings identical.
Under both conditions, keypoint trajectories consistently outperform 6D rot+trans~\cite{peng2023hoi} and 9D rotation-matrix+trans~\cite{li2023controllable}, achieving the lowest jerk, translation error, and orientation error.
As shown in \Fref{fig:rotation_3verts}, the advantage is particularly evident for articulated objects.

\begin{table}[t]
\renewcommand{\arraystretch}{1.2}
\begin{center}
    \resizebox{\linewidth}{!}{\begin{tabular}{l |cc |cc |cc}
       \toprule
         % \cmidrule{2-13}
        % \hline
           \multirow{2}{*}{Design Choice}  & \multicolumn{2}{c|}{ParaHome (2~obj.)} & \multicolumn{2}{c|}{ParaHome (3~obj.)} & \multicolumn{2}{c}{OMOMO} \\
           \cmidrule{2-3}\cmidrule{4-5}\cmidrule{6-7}
            & $C_{acc}^{tem}  \uparrow$ & $C_{acc}^{body} \uparrow$  & $C_{acc}^{tem}  \uparrow$ & $C_{acc}^{body} \uparrow$ & $C_{acc}^{tem}  \uparrow$ & $C_{acc}^{body} \uparrow$\\
        \midrule
        Euclidean Distance~\cite{xue2025guiding} & 0.612 & 0.857 & 0.642 & 0.854 & 0.848 & 0.858 \\
        Binary Label  & 0.565 & 0.802 & 0.594 & 0.806 & 0.795 & 0.804 \\
        Contact Pairs~\cite{ron2025hoidini} & 0.589 & 0.821 & 0.612 & 0.813 & 0.808 & 0.815 \\
        Ours & \textbf{ 0.653}  & \textbf{0.902} &  \textbf{ 0.683}  & \textbf{0.901} & \textbf{ 0.893}  & \textbf{0.902}   \\

        \bottomrule
\end{tabular}}
\caption{\textbf{Ablation results on the impact of different contact representations on ParaHome (2--3 objects) and OMOMO.} Best results are highlighted in \textbf{bold}.}

\label{table:contact_type}
\end{center}
\vspace{-28pt}
\end{table}
\myparagraph{Effect of Contact Representations.}
% We compare four types of intermediate contact supervision.
As shown in \Tref{table:contact_type}, the proposed distance field achieves the best results across all metrics on ParaHome and OMOMO, confirming that it provides a more informative and learnable interaction cue than the alternatives.
Binary labels suffer from extreme sparsity, resulting in weak gradients that make it difficult for the model to learn fine-grained contact transitions.
Raw Euclidean distances, are absolute and unnormalized, increasing learning complexity and sensitivity to variations in object size.
Contact Pairs~\cite{ron2025hoidini} encode hand-object contact through a fixed number of selected contact locations on the object surface, resulting in relatively sparse contact cues.

\subsection{Limitations and Discussions}
\label{sec:limitations}
The noise optimization in Stage III adds computational overhead (approximately 4 minutes per 124-frame sequence), which limits real-time deployment; developing feed-forward alternatives for contact refinement is a promising direction.
Physical artifacts such as foot floating and minor penetrations remain present in our results, as they do across existing HOI generation methods, given the fundamental challenges of the HOI generation task; integrating physics-based constraints or learned collision handling could help mitigate these in future work.
\section{Conclusion}
% We introduced a unified point-based representation for human--object interaction generation, modeling both humans and objects as trajectories of surface points in a shared Euclidean space. This formulation naturally supports interactions with a variable number of objects and articulated objects without requiring explicit joint-type specification, while preserving surface geometry for physically plausible contacts. To tackle the combinatorial complexity of multi-object HOIs, we further factorize generation into three sequential stages: object motion generation, correspondence-aware distance field prediction, and whole-body motion synthesis with contact optimization. Experiments on ParaHome and OMOMO demonstrate state-of-the-art motion quality and physical plausibility, especially in challenging multi-object and articulated scenarios.
We presented surface keypoint trajectories as an object motion representation for human-object interaction generation that handles a variable number of rigid and articulated objects without explicit joint-type specification.
We also introduced a spatio-temporal contact distance field that extends distance-based contact modeling to whole-body, multi-object, and articulated settings, predicted as a standalone intermediate representation in a factorized three-stage pipeline.
Experiments on ParaHome, HIMO, ARCTIC, and OMOMO demonstrate better or comparable performance to existing methods across single-object, multi-object, and articulated interaction settings.

% Bibliography
\bibliographystyle{ACM-Reference-Format}
\bibliography{main}

\newpage

\begin{figure*}
    \centering
    \includegraphics[width=1.0\linewidth]{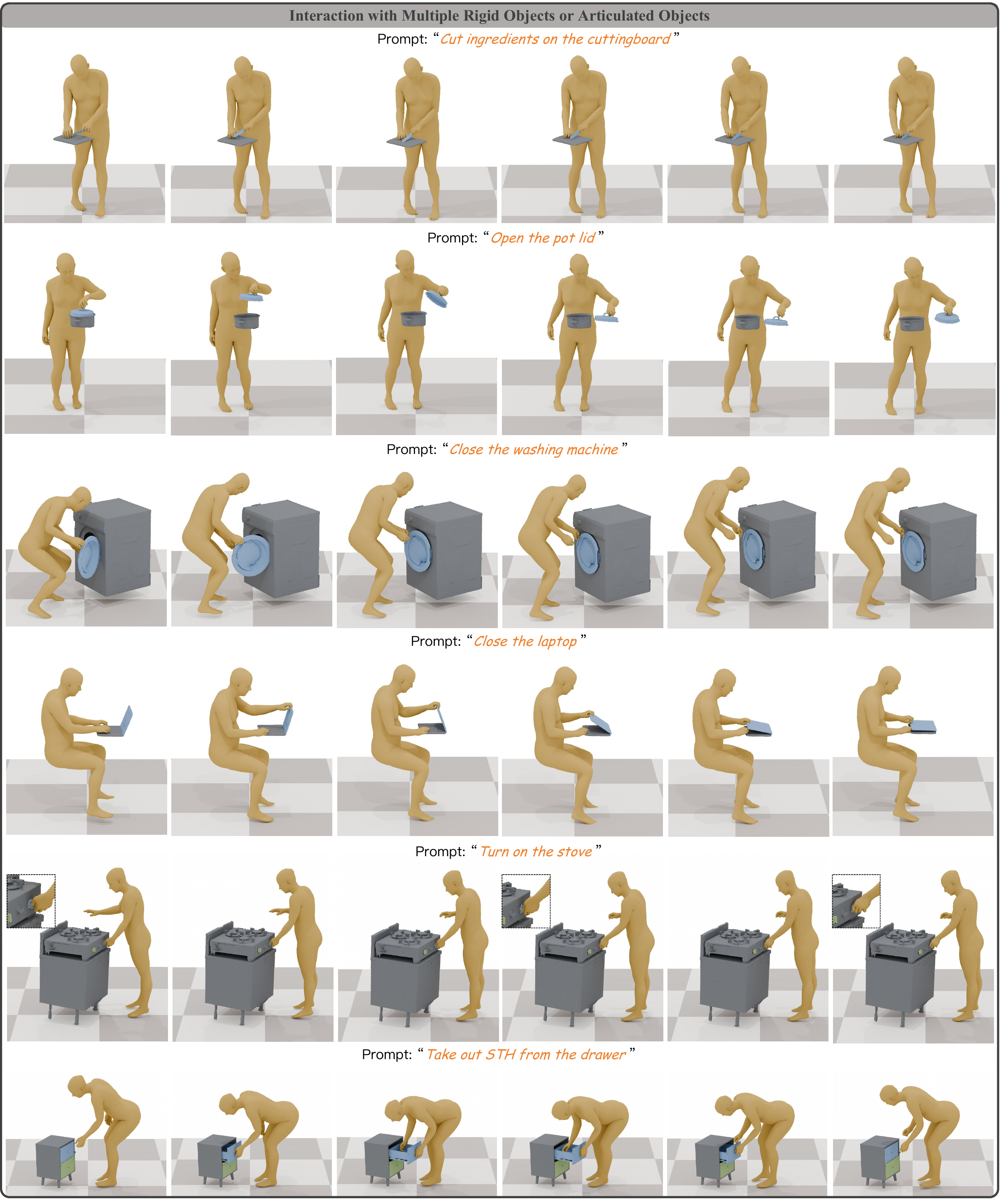}
    \vspace{-2em}
    \caption{\textbf{Results Gallery.} We provide additional results on multi-object and articulated HOI generation scenarios on the ParaHome dataset.}
    \label{fig:more_vis}
\end{figure*}

\clearpage
\appendix
\setcounter{page}{1}
\setcounter{table}{0}
\setcounter{figure}{0}

We present more information in the supplementary material, organized as follows:
\begin{itemize}[itemsep=1pt,topsep=0pt,leftmargin=18pt]
\item \ref{supp:baselines}: Implementation Details of Baseline Methods.
\item \ref{supp:metrics}: Details of Metrics and Evaluation.
\item \ref{supp:diffusion}: Flow Matching Preliminary.
\item \ref{supp:implementation}: Implementation Details of Our Method.
\item \ref{supp:additional_results}: Additional Results.
% \item \ref{supp:qualitative}: Additional Qualitative Results.
\item \ref{supp:user_study}: Details of User Study.
\item \ref{supp:limitations}: Failure Cases.
\item \ref{supp:video}: Supplemental Video.
\end{itemize}

\section{Implementation Details of Baseline Methods}
\label{supp:baselines}
\myparagraph{HIMO-Gen~\citep{lv2024himo}}: HIMO is the most similar work to ours. We train separate models for interactions with two and three objects, respectively, since it does not support a variable number of objects.
% conceptually close to our task, we follow it to test on two and three objects for interaction, denoted as HIMO${2o}$ and HIMO${3o}$. 
We retrain it using its original input representation, which includes full-body global joints and SMPL-X parameters.

\myparagraph{HOI-Diff~\citep{peng2023hoi}}: We use the original human pose representation from HOI-Diff in the HumanML3D~\cite{Guo_2022_CVPR} format, which contains 263 dimensions without hand motion modeling. We retrain the model using this original representation and convert the generated outputs to global joint coordinates and then obtain the marker motion from the recovery SMPL-X mesh vertices under our proposed metrics.

\myparagraph{CHOIS~\citep{li2023controllable}}: CHOIS was originally conditioned on both text and waypoint trajectories. To enable a fair comparison under text-only settings, we remove the waypoint input and adapt the input dimensions accordingly.

\myparagraph{ROG~\citep{xue2025guiding}}:
ROG is a recent open-source baseline for single-object HOI generation. It represents object geometry using boundary-focused sparse keypoints and constructs an interactive distance field to model human-object relations. We retrain their model and evaluate it on the OMOMO benchmark.

\myparagraph{HOIDiNi~\citep{ron2025hoidini}}:
HOIDiNi is a recent text-driven HOI generation method that optimizes the diffusion noise space to improve contact accuracy and motion plausibility. We retrain their model and evaluate it on the OMOMO benchmark.

\section{Details of Metrics and Evaluation}
\label{supp:metrics}
For detailed information regarding metrics employed in human motion generation, including \textbf{FID}, \textbf{R-Precision}, and \textbf{Diversity}, we refer readers to \cite{tevet2023human, Guo_2022_CVPR} for a comprehensive understanding. We first build an evaluator by following the architecture of the widely used motion--text evaluator~\cite{Guo_2022_CVPR}, which consists of a convolutional movement encoder, a GRU~\cite{gru}-based motion encoder, and a GRU-based text encoder using GloVe~\cite{glove} embeddings. 
The evaluator is trained on our selected marker-based motion representation, which captures the key dynamics that are most relevant to HOI generation.

\myparagraph{Jerk$_{obj}$.}  
It quantifies object motion smoothness, defined as the rate of change of acceleration.
Given a sequence of object keypoints trajectory $\mathbf{O}_{1:T}$ with $T$ frames, we compute:
\begin{equation}
\mathrm{Jerk}
=\frac{1}{T-3}\sum_{t=1}^{T-3}\left\|\mathbf{a}_{t+1}-\mathbf{a}_{t}\right\|_2,
\label{eq:jerk}
\end{equation}
where $\mathbf{a}_t$ denotes the per-frame acceleration of the object keypoints obtained by applying finite differences to $\mathbf{O}_{1:T}$, and $\|\cdot\|_2$ is taken over the concatenated keypoint coordinates (in centimeters).
Lower jerk scores indicate smoother motion.

\myparagraph{Temporal Contact Accuracy $C_{acc}^{tem}$ and Body Contact Accuracy $C_{acc}^{body}$.} \paragraph{Contact definition.}
For each sequence with $N_{\mathrm{obj}}$ active rigid components ($N_{\mathrm{obj}}\le N$), we
evaluate contact \emph{per component} and average over components. Let
$\mathbf{H}_t\in\mathbb{R}^{M\times3}$ be the full set of $M$ human markers and
$\mathbf{S}^{o}_t\in\mathbb{R}^{Q\times3}$ the surface points of component $o$ at time $t$. For each
component we compute the per-marker minimum distance to that component's surface,
\begin{equation}
\mathbf{d}^{o}_t \;=\; \min_{i\in\{1,\ldots,Q\}}\mathtt{dist}\!\big(\mathbf{H}_t,\,\mathbf{S}^{o}_t[i]\big)
\;\in\;\mathbb{R}^{M},
\end{equation}
and a binary contact label using a distance threshold $\tau=2\,\text{cm}$,
\begin{equation}
\mathbf{g}^{o}_t \;=\; \mathbb{I}\!\left[\mathbf{d}^{o}_t < \tau\right]\;\in\;\{0,1\}^{M}.
\end{equation}
We apply the same procedure to the ground-truth and predicted trajectories to obtain
$\mathbf{g}^{o,\text{gt}}_t$ and $\mathbf{g}^{o,\text{pred}}_t$. Contact accuracy is then computed
per component from these labels and averaged over the $N_{\mathrm{obj}}$ components (and over
sequences). We report a body-level score over all $(t,m)$ pairs and a temporal (frame-level) score
in which a frame is in contact if any marker is, i.e.\ $\bigvee_{m}\mathbf{g}^{o}_t[m]$.

% \begin{equation}
% d_{t,m} \;=\; \min_{i \in \{1,\ldots,V\}} \big\| \mathbf{h}_{t,m} - \mathbf{s}_{t,i} \big\|_2 .
% \end{equation}
% \begin{equation}
% c_{t,m} \;=\; \mathbb{I}\!\left[d_{t,m} < \tau\right],
% \end{equation}

\paragraph{(A) Temporal Contact Accuracy.}
We first define a frame-level contact indicator that is active if \emph{any} marker is in contact:
% \begin{equation}
% C_t \;=\; \bigvee_{m=1}^{M} c_{t,m}.
% \end{equation}
\begin{equation}
G_t \;=\; \mathbb{I}\!\left[\mathbf{1}^\top \mathbf{g}_t > 0\right].
\end{equation}
Let $G_t^{\text{gt}}$ and $G_t^{\text{pred}}$ denote ground-truth and predicted frame-level contact labels.
We compute the confusion counts:
\begin{align}
\mathrm{TP} &= \sum_{t=1}^{T} \mathbb{I}\!\left[G_t^{\text{gt}}=1 \text{ }\&\text{ } G_t^{\text{pred}}=1\right], \\
\mathrm{FP} &= \sum_{t=1}^{T} \mathbb{I}\!\left[G_t^{\text{gt}}=0 \text{ }\&\text{ } G_t^{\text{pred}}=1\right], \\
\mathrm{TN} &= \sum_{t=1}^{T} \mathbb{I}\!\left[G_t^{\text{gt}}=0 \text{ }\&\text{ } G_t^{\text{pred}}=0\right], \\
\mathrm{FN} &= \sum_{t=1}^{T} \mathbb{I}\!\left[G_t^{\text{gt}}=1 \text{ }\&\text{ } G_t^{\text{pred}}=0\right].
\end{align}
Then the temporal contact accuracy is:
\begin{align}
C_{acc}^{tem} &= \frac{\mathrm{TP}+\mathrm{TN}}{\mathrm{TP}+\mathrm{FP}+\mathrm{TN}+\mathrm{FN}}.
\end{align}

\paragraph{(B) Body Contact Accuracy.}
We treat each marker-time pair $(t,m)$ as an independent binary classification target.
Let $g_{t,m}^{\text{gt}}$ and $g_{t,m}^{\text{pred}}$ denote ground-truth and predicted marker-level contact labels (i.e., the $m$-th entries of $\mathbf{g}_t^{\text{gt}}$ and $\mathbf{g}_t^{\text{pred}}$).
We compute the confusion counts over all $(t,m)$:
\begin{align}
\mathrm{TP}_m &= \sum_{t=1}^{T}\sum_{m=1}^{M} 
\mathbb{I}\!\left[g_{t,m}^{\text{gt}}=1 \text{ }\&\text{ } g_{t,m}^{\text{pred}}=1\right], \\
\mathrm{FP}_m &= \sum_{t=1}^{T}\sum_{m=1}^{M} 
\mathbb{I}\!\left[g_{t,m}^{\text{gt}}=0 \text{ }\&\text{ } g_{t,m}^{\text{pred}}=1\right], \\
\mathrm{TN}_m &= \sum_{t=1}^{T}\sum_{m=1}^{M} 
\mathbb{I}\!\left[g_{t,m}^{\text{gt}}=0 \text{ }\&\text{ } g_{t,m}^{\text{pred}}=0\right], \\
\mathrm{FN}_m &= \sum_{t=1}^{T}\sum_{m=1}^{M} 
\mathbb{I}\!\left[g_{t,m}^{\text{gt}}=1 \text{ }\&\text{ } g_{t,m}^{\text{pred}}=0\right].
\end{align}
The body contact accuracy is then defined as:
\begin{align}
C_{acc}^{body} &= \frac{\mathrm{TP}_m+\mathrm{TN}_m}{\mathrm{TP}_m+\mathrm{FP}_m+\mathrm{TN}_m+\mathrm{FN}_m}.
\end{align}

\myparagraph{Penetration Score.}
Following~\citet{li2023controllable}, we quantify body-object interpenetration by querying the precomputed signed distance field at each marker location.
At time step $t$, $\mathrm{SDF}_t(\cdot)$ returns the signed distance to the object surface, with negative values indicating penetration.
When multiple objects are present, $\mathrm{SDF}_t(\cdot)$ takes the minimum signed distance over all active objects.
We compute:
\begin{equation}
\text{Pene} \;=\; \frac{1}{TM}\sum_{t=1}^{T}\sum_{m=1}^{M} \max\bigl(-\mathrm{SDF}_t\!\left(\mathbf{H}_{t}(m)\right),\, 0\bigr),
\end{equation}
measured in centimeters.

\section{Flow Matching Preliminaries.}
\label{supp:diffusion}
In each stage, we use a diffusion-based generative model, implemented via rectified flow~\cite{liu2022flow} and ODE sampling. Given a condition $c$ (e.g., text and geometry features), we model the generative process as a continuous-time flow defined by an ODE:
\begin{equation}
\frac{d\mathbf{x}(t)}{dt} = \mathbf{v}_\theta(\mathbf{x}(t), t, c), \quad t\in[0,1],
\label{eq:fm_ode}
\end{equation}
where $\mathbf{v}_\theta$ is a neural velocity field.

\noindent\textbf{Training objective.}
We construct an interpolation path between data $\mathbf{x}_0 \sim p_{\text{data}}$ and noise $\mathbf{x}_1 \sim \mathcal{N}(\mathbf{0},\mathbf{I})$:
\begin{equation}
\mathbf{x}_t = (1-t)\mathbf{x}_0 + t\mathbf{x}_1.
\label{eq:fm_interp}
\end{equation}
The corresponding target velocity is
\begin{equation}
\mathbf{v}^\ast(\mathbf{x}_t,t) = \frac{d\mathbf{x}_t}{dt} = \mathbf{x}_1-\mathbf{x}_0.
\label{eq:fm_target_v}
\end{equation}
We train $\mathbf{v}_\theta$ via flow matching:
\begin{equation}
\mathcal{L}_{\text{FM}}
=\mathbb{E}_{\mathbf{x}_0,\mathbf{x}_1,t}\Big[
\big\|
\mathbf{v}_\theta(\mathbf{x}_t,t,c) - \mathbf{v}^\ast(\mathbf{x}_t,t)
\big\|_2^2
\Big].
\label{eq:fm_loss}
\end{equation}

\noindent\textbf{Sampling.}
At inference time, we sample $\mathbf{x}(1)\sim\mathcal{N}(\mathbf{0},\mathbf{I})$ and solve the ODE in Eq.~\eqref{eq:fm_ode} backward from $t{=}1$ to $t{=}0$ using an ODE solver (e.g., Euler or Heun):
\begin{equation}
\mathbf{x}(t-\Delta t)=\mathbf{x}(t)-\Delta t\cdot \mathbf{v}_\theta(\mathbf{x}(t),t,c).
\label{eq:fm_solver}
\end{equation}
The final sample $\mathbf{x}(0)$ is the generated sequence.

\begin{figure}[t]
    \centering
    \includegraphics[trim={0 0.3cm 0 0},clip, width=0.8\linewidth]{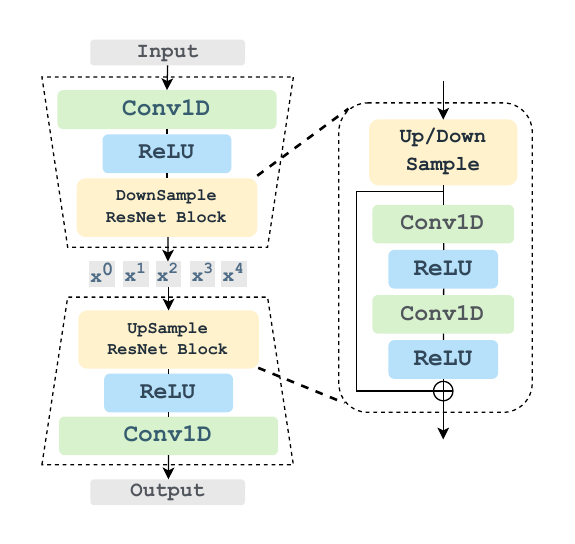}
    \caption{Architecture of the Causal AutoEncoder.}
    \label{fig:autoencoder}
\end{figure}

\section{Implementation Details of Our Method.}
\label{supp:implementation}
\begin{figure}[h]
    \centering
    \includegraphics[width=1.0\linewidth]{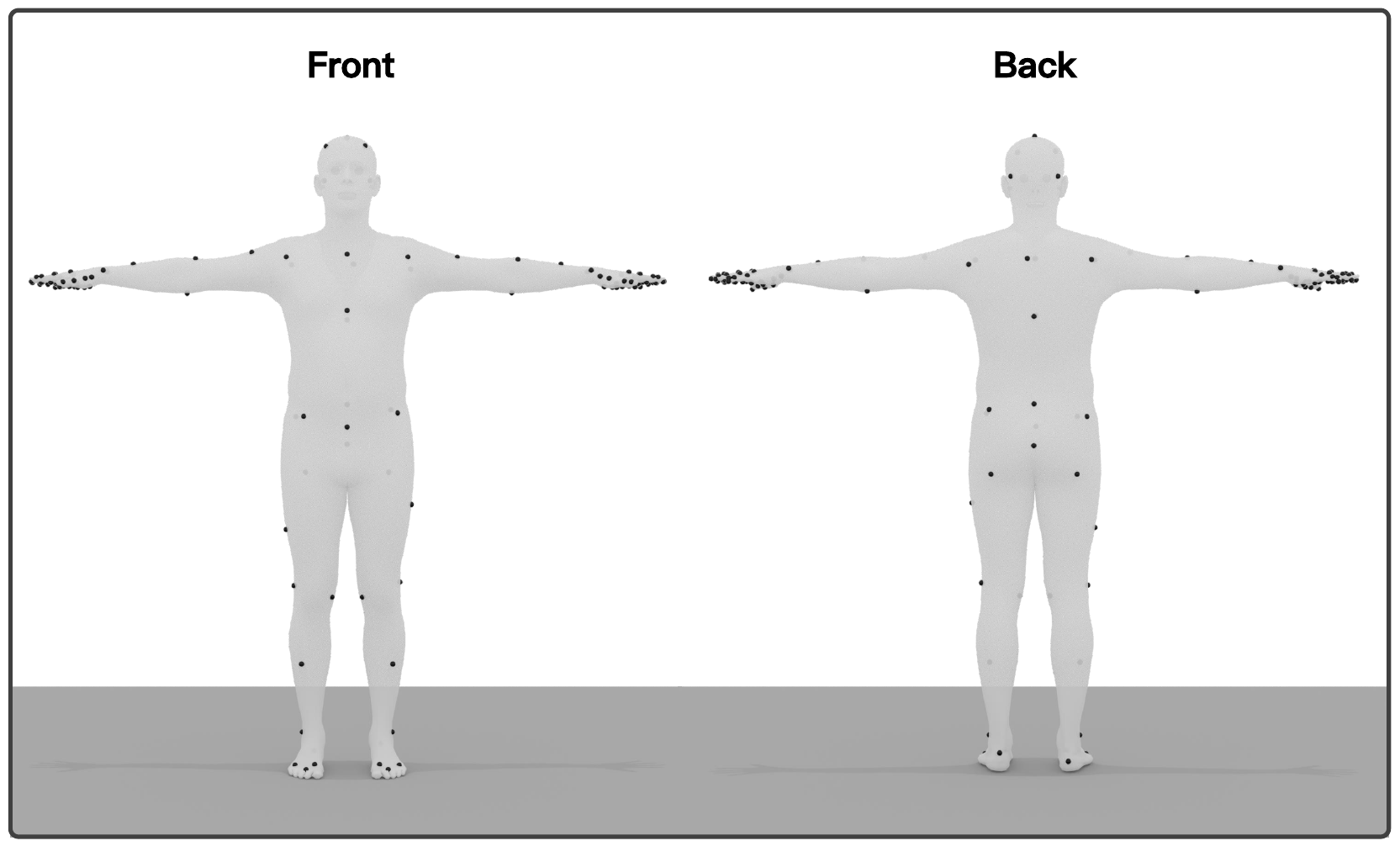}
    \caption{Illustration of our human marker representation.}
    \label{fig:markers}
\end{figure}

\myparagraph{Human Marker Representation.}
As illustrated in~\Fref{fig:markers}, we select a total of 138 surface markers covering the full human body. 
We place denser markers on the hands, since fine-grained contacts often occur on the palms and fingers. 
Specifically, we use 38 markers on the main body, 40 markers on each hand, 6 markers on each foot, 5 markers on the head, and 3 markers on the hip/buttocks region.

\myparagraph{Object Representation.} For keypoint selection, we sample $K$ local points from the vertex set using Farthest Point Sampling (FPS). For the surface representation used in Stages~II and~III, we sample $Q=384$ surface points via Poisson-disk sampling to obtain approximately uniform coverage of the object surface, which also reduces the computational cost of distance-field computation.

\myparagraph{Main Model Architecture Details}. We use a CLIP text encoder to obtain prompt embeddings and an MLP to encode object geometry represented by BPS features or sparse object waypoints if available. 
Across all three stages, the input sequences, including object motion, contact distance fields, and human marker motion, are compressed into a 64-dimensional latent space using task-specific autoencoders with the same architecture as shown in the \fref{fig:autoencoder}. The AutoEncoder is a 3-layer ResNet-based encoder-decoder architecture with a total downsampling rate of 4. For each diffusion backbone, we utilize an 8-layer AdaLN-Zero transformer~\cite{dit} encoder with a hidden dimension of 1024 and 4 heads. We also adopt several recent architectural advances for better sequence modeling: Rotary Positional Embeddings (RoPE)~\cite{su2024roformer} and QK Normalization~\cite{henry2020query} are employed within attention layers, and SwiGLU activations~\cite{shazeer2020glu} are used in the feed-forward networks (FFNs).

\myparagraph{Causal Autoencoder}
\label{sec:supp_ae}
All three stages operate in a compact latent space produced by a \emph{causal}
temporal autoencoder (Fig.~1). Rather than one joint encoder, we use three
structurally identical but independently trained branches---one each for object
motion, the contact distance field, and human marker motion---so every modality
has a dedicated codec while sharing the same design. Each branch maps a length-$T$
per-frame feature sequence to a latent of length $T/4$ with $d{=}64$ channels; the
Stage~I--III models predict flow-matching velocities directly on these latents.

\paragraph{Causal temporal convolutions.}
Every convolution in the encoder and decoder is \emph{left-padded only}, so the
latent at temporal block $\tau$ summarizes exclusively input frames $t\le 4\tau$
and never accesses the future. This makes the encoding of the clean initial state
($\mathbf{H}_{\text{init}},\mathbf{O}_{\text{init}}$) and of the generated
continuation consistent under the same operator, which is what our
initial-state conditioning (Sec.~3.2) relies on.

\paragraph{Architecture.}
The encoder lifts the input to a hidden width of $512$ with a $1$D convolution
($k{=}3$), then applies two downsampling stages; each stage is a strided causal
convolution ($k{=}4$, stride $2$) that halves the temporal length, followed by
three residual blocks. The two stages give the total temporal compression of
$4\times$. Each residual block is a pre-activation unit with SiLU (swish) gating,
a dilated causal convolution ($k{=}3$; dilations $9,3,1$ across the three blocks
to enlarge the temporal receptive field), a $1{\times}1$ convolution, and dropout
$0.2$. A final convolution maps the hidden state to the $d{=}64$ latent. The
decoder mirrors this: it expands the latent to width $512$, applies two
upsampling stages (nearest-neighbor $\times 2$ followed by a causal convolution),
each preceded by three residual blocks, and two output convolutions map back to
the input dimension.

\paragraph{Per-modality inputs and latents.}
\emph{Human.} The $M{=}138$ surface markers form a $3M{=}414$-d per-frame vector,
encoded into one latent stream $\mathbf{z}^{H}\!\in\!\mathbb{R}^{(T/4)\times d}$.
\emph{Object.} Each rigid component is described by its $K{=}3$ keypoints
($3K{=}9$-d per frame) and is encoded \emph{independently} by the shared object
branch, so an interaction with $N$ components yields $N$ latent streams
$\mathbf{z}^{O}\!\in\!\mathbb{R}^{N\times(T/4)\times d}$; weight sharing lets a
single model handle a variable number of objects.
\emph{Contact.} The contact distance field $\mathbf{D}\!\in\!\mathbb{R}^{T\times
M_c\times Q}$ (per component) is compressed \emph{per marker}: for each of the
$M_c{=}47$ selected markers, its length-$T$, $Q{=}384$-d proximity trajectory is
encoded independently, giving $\mathbf{z}^{D}\!\in\!\mathbb{R}^{N\times M_c\times
(T/4)\times d}$. Encoding each marker independently preserves the sparse,
high-frequency structure of the field (best reconstruction and contact recall)
and keeps the codec agnostic to the number of markers and components.

\paragraph{Training objective.}
Each branch is trained with a reconstruction loss over valid frames only (padded
frames excluded). Human and object motion use the Smooth-$L_1$ loss, with the
object loss additionally masked to the real components. Since $\mathbf{D}$ is
dominated by a near-zero background, a uniform loss under-weights the rare
in-contact entries; we therefore up-weight cells with $\mathbf{D}>0.5$ by
$(1+\lambda_c)$, which substantially improves contact recall. The three branches
have disjoint parameters and are optimized independently, so the contact codec can
be trained separately and combined with the motion codec.

\paragraph{Latent standardization.}
Before the Stage~II contact diffusion, we standardize the contact latent to zero
mean and unit variance using statistics precomputed on the training set, and
invert this after sampling; this stabilizes flow matching on the contact latent,
whose raw scale differs from the motion latents.

\paragraph{Optimization.}
Each autoencoder is trained on ParaHome and OMOMO with $64$-frame clips, batch
size $128$, and AdamW ($\beta_1{=}0.9,\beta_2{=}0.99$) at learning rate
$2\times10^{-4}$.

\myparagraph{Conditioning.}
We condition the model on the initial human state and initial object state by temporally concatenating them with the input human sequence and object sequence, respectively. 
The BPS embedding is repeated across $T + L_{init}$ time steps and fused with the object motion via feature-wise concatenation. 
When sparse object waypoints is available,  we pad it to length $T + L_{init}$, and project it to the latent feature dimension, and then fuse with the object motion through feature-wise concatenation. 
When full object trajectories are provided as additional conditions (Stage~III), we incorporate them by temporally concatenating the trajectories with the input human motion.

\myparagraph{Training.} The model is implemented in PyTorch~\citep{paszke2019pytorch} and trained on a single NVIDIA A6000 GPU. Our training setting involves 200k steps and utilize a batch size of 128 and employ the AdamW optimizer~\citep{loshchilov2017decoupled} with a learning rate set at $2 \times 10^{-4}$.

During training, we use the AdamW optimizer with $\beta_1{=}0.9$ and $\beta_2{=}0.99$. 
Following prior work, we train the autoencoders on ParaHome and OMOMO with a batch size of 128, where each sample contains 64 frames. We then train the diffusion models for 200K steps with a batch size of 64 and a maximum sequence length of 124 frames. The learning rate is set to $2\times10^{-4}$.

For additional experiments on the ARCTIC and HIMO benchmarks, we follow their task settings using our model implementation.

\begin{figure*}
    \centering
    \begin{overpic}[width=0.95\linewidth]{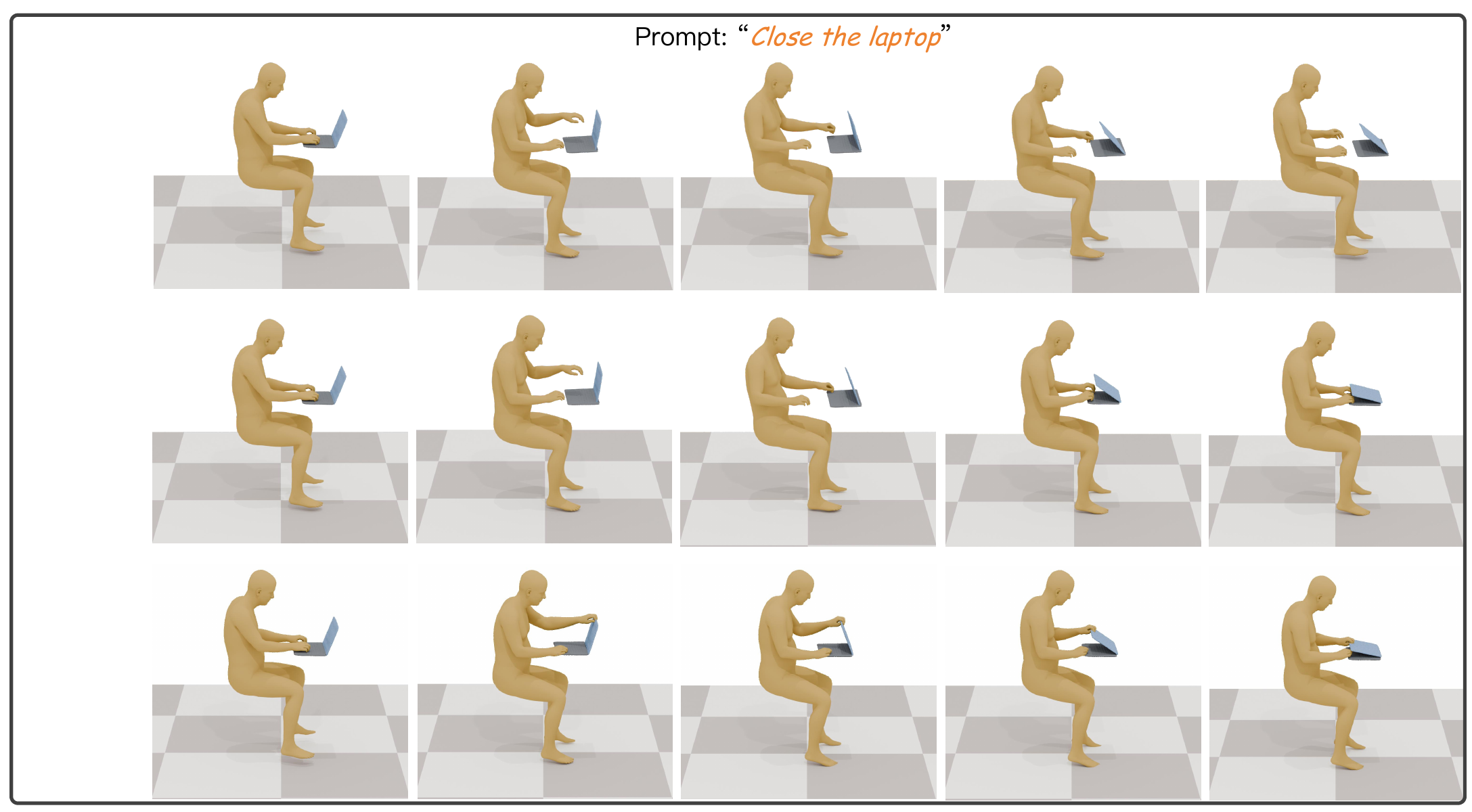}
     \put(6,120){\small{w/o Contact}}
      \put(6,110){\small{Optimization}}
     \put(6,210){\small{w/o Separate}}
      \put(15,200){\small{Stage}}
     \put(6,30){\small{Ours (Full)}}
    \end{overpic}
    \vspace{-1em}
\caption{\textbf{Visual Results of Ablation Study.} }
    \label{fig:ablation_vis}
\end{figure*}

\begin{figure*}[t]
    \centering
    \begin{overpic}[width=0.92\linewidth]{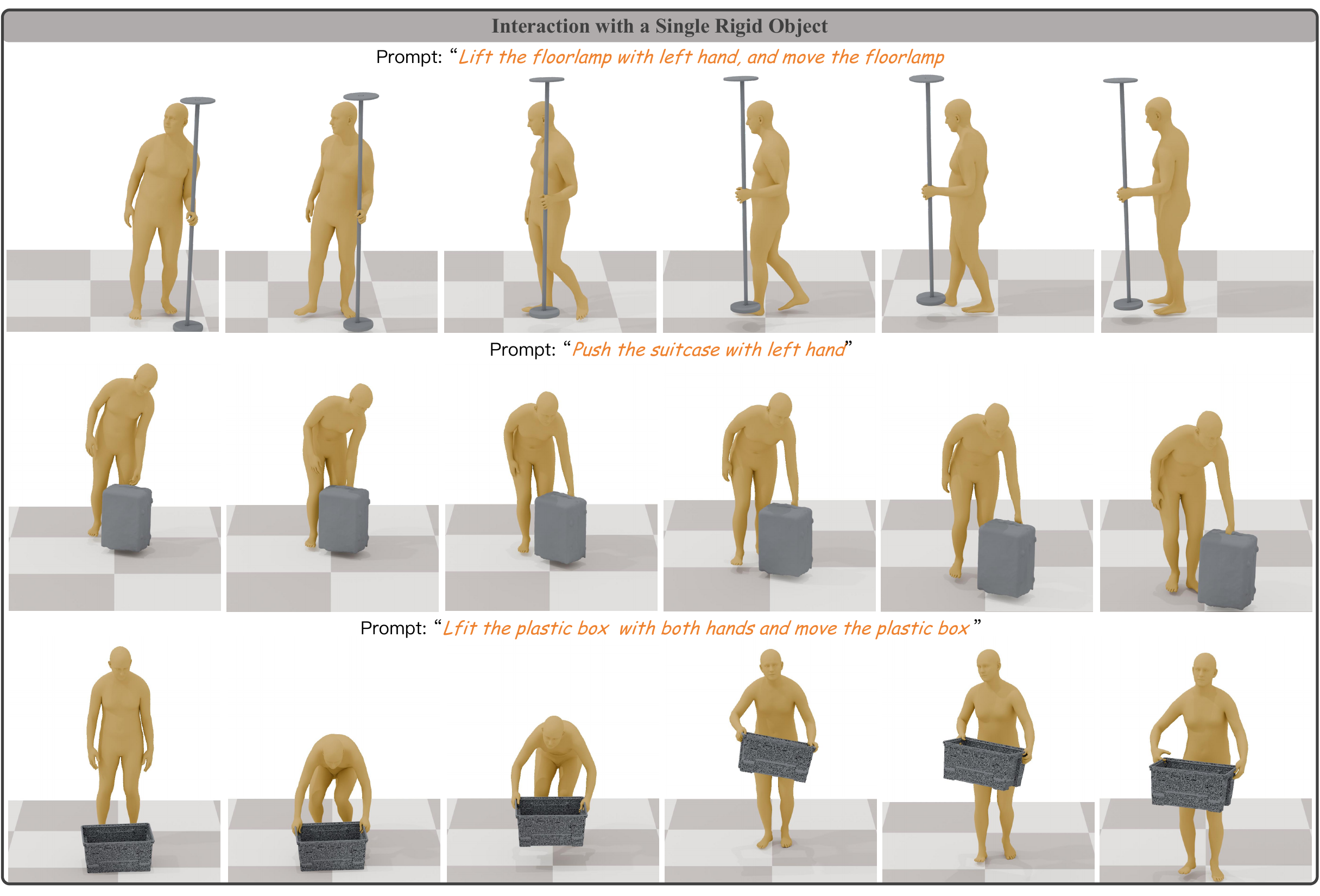}
     % \put(9,120){\small{w/o Contact}}
     %  \put(9,110){\small{Optimization}}
     % \put(9,210){\small{w/o Separate}}
     %  \put(16,200){\small{Stage}}
     % \put(9,30){\small{Ours (Full)}}
    \end{overpic}
    % \vspace{-1.6em}
\caption{\textbf{Results Gallery.} We provide additional results for single-object interaction on OMOMO dataset}
% \vspace{-1em}
    \label{fig:more_vis_omomo}
\end{figure*}

\begin{table*}[t]
\renewcommand{\arraystretch}{1.0}
\begin{center}
    \resizebox{1.0\linewidth}{!}{\begin{tabular}{l | l |ccccc | ccc| ccccc | ccc}
       \toprule
        & & \multicolumn{8}{c|}{2 objects/components} & \multicolumn{8}{c}{3 objects/components} \\
        \cmidrule{3-10}\cmidrule{11-18}
           Dataset & Method& \multicolumn{5}{c|}{Motion} & \multicolumn{3}{c|}{Interaction} &\multicolumn{5}{c|}{Motion} & \multicolumn{3}{c}{Interaction}\\
        \cmidrule{3-7}\cmidrule{8-10}\cmidrule{11-15}\cmidrule{16-18}
          & \ &  FID $\downarrow$  & $R_{prec}$ $\uparrow$ & Div $\rightarrow$ & FS $\downarrow$ & Jerk$_{obj}$ $\downarrow$ & $C_{acc}^{tem}\uparrow$ & $C_{acc}^{body}$$\uparrow$ & Pene $\downarrow$  &  FID $\downarrow$  & $R_{prec}$ $\uparrow$ & Div $\rightarrow$ & FS $\downarrow$ & Jerk$_{obj}$ $\downarrow$ & $C_{acc}^{tem}\uparrow$ & $C_{acc}^{body}$$\uparrow$ & Pene $\downarrow$ \\
        \midrule
        \multirow{3}{*}{\rotatebox[origin=c]{45}{ParaHome}}
          & Real motion (reference)             &  0.00  & 0.727  & 7.78   & 0.0039 & 0.15 & - &  -    & -                            & 0.00    &  0.679  & 7.47  & 0.0031  & 0.05     & -     &  -   &- \\
         & w/o Separate Stage    &4.52    & 0.673  & 7.69  & \textbf{0.0027}  & 1.14 & 0.544  &  0.861 & 0.550 & 6.60 & 0.580  &7.30 &\textbf{0.0017}          & 2.34    &  0.646 &0.898 & 0.773\\
        & w/o Contact Optimization              &  4.55  & 0.691 & \textbf{7.85}    & 0.0064 & \textbf{0.72} & 0.613 & 0.883 & 0.619              & 6.14           &  0.581  & 7.90  & 0.0022  & \textbf{0.40}  & 0.657 & \textbf{0.911} & \textbf{0.763}\\
          & Ours   & \textbf{4.49}    &\textbf{0.707}  & 8.38 & 0.0035 & \textbf{0.72} &\textbf{0.669}& \textbf{0.896}  &  \textbf{0.536}             & \textbf{6.09} & \textbf{0.598} & \textbf{7.57} & \textbf{0.0017}     & \textbf{0.40}   & \textbf{0.680}   &  0.906  & 0.776\\
        \midrule
        \multirow{3}{*}{\rotatebox[origin=c]{45}{HIMO}}
          & Real motion  (reference)            &  0.000  & 0.729  & 11.905   & 0.0007 & 0.08 & - &  -    & -                            & 0.267    &  0.713  & 9.755  & 0.0004  & 0.10     & -     &  -   &- \\
         & w/o Separate Stage    &7.821    & 0.579  & \textbf{10.963}  & 0.0017  & 0.98 & 0.674  &  0.829 & 0.535 & 3.571 & 0.547  &10.212 &0.0026          & 1.44    &  0.693 & 0.812 & 0.638\\
        & w/o Contact Optimization              &  6.893  & 0.608 & 10.358    & 0.0012 & \textbf{0.17} & 0.706 & 0.897 & 0.572              & 2.345           &  0.613  & 9.907  & 0.0022  & \textbf{0.19}  & 0.737 & 0.879 & \textbf{0.603}\\

          & Ours   & \textbf{6.572}    &\textbf{0.612}  & 11.021 & \textbf{0.0011} & \textbf{0.17} & \textbf{0.731} & \textbf{0.912}  &  \textbf{0.519}       & \textbf{2.273} & \textbf{0.628} & \textbf{9.812} &\textbf{0.0015}     &\textbf{0.19}   & \textbf{0.751}   &  \textbf{0.883}  & 0.621\\
        \bottomrule
\end{tabular}}
\caption{\textbf{Ablation results on ParaHome and HIMO for multi-object and articulated HOI generation.} We compare against our full method and its variants. Best results are highlighted in \textbf{bold}. 
}
\label{table:ablation_main}
\end{center}
\vspace{-2.5em}
\end{table*}

\myparagraph{Contact Optimization Detail.}
During inference in Stage~III, we perform noise-space optimization with an ODE solver using 5 denoising steps and 200 noise optimization iterations. We use a cosine-decayed learning rate with an initial value of $0.05$. The optimization objective is a weighted sum of losses, with $\lambda_{\text{pene}}=0.001$.

\myparagraph{Marker-to-SMPLX Fitting Model.}
For rendering and evaluation, we train a lightweight 10-layer ResNet for 100 epochs on ParaHome and OMOMO to regress SMPL-X parameters from the input human markers. 
We use the neutral-gender SMPL-X model and predict the shape coefficients $\boldsymbol{\beta}$, SMPL-X pose $\boldsymbol{\theta}$ and global translation. During inference, we further perform a 200-step test-time optimization to refine the fitted parameters and better align the recovered mesh with the marker observations. Specifically, given a set of target markers $\mathbf{H}\in\mathbb{R}^{T\times M\times 3}$, we only optimize $\boldsymbol{\theta}$ and shape coefficients $\boldsymbol{\beta}$ to minimize the L1 reconstruction error between SMPL-X markers and target markers.
We optimize body poses, hands poses, and shape coefficients ($\boldsymbol{\beta}$) with separate Adam optimizers at a learning rate of $10^{-3}$.
We run 400 optimization steps and compute the marker loss:
\begin{equation}
\mathcal{L}_{\text{marker}}=
\left\|
\mathbf{H} - \tilde{\mathbf{H}}(\boldsymbol{\theta},\boldsymbol{\beta})
\right\|_{1},
\end{equation}
where $\tilde{\mathbf{H}}(\cdot)$ denotes the SMPL-X markers extracted from the reconstructed mesh vertices.
During fitting, facial and eye pose parameters are fixed to zero.
After optimization, we use the fitted SMPL-X parameters to recover the full human mesh for visualization and metric computation.

\Tref{tab:smplx_fit} reports the fitting error before and after this optimization: the feed-forward regressor provides a coarse initialization, which the optimization refines to a marker error of \textbf{21.88}\,mm, confirming that the recovered mesh closely matches the input markers and does not introduce a significant bottleneck in the reported human-motion metrics.

\section{Additional Results}
\label{supp:additional_results}

\begin{figure*}[t]
% \vspace{-2em}
    \centering
     \begin{overpic}[trim={0 0 0 0}, width=0.88\linewidth]{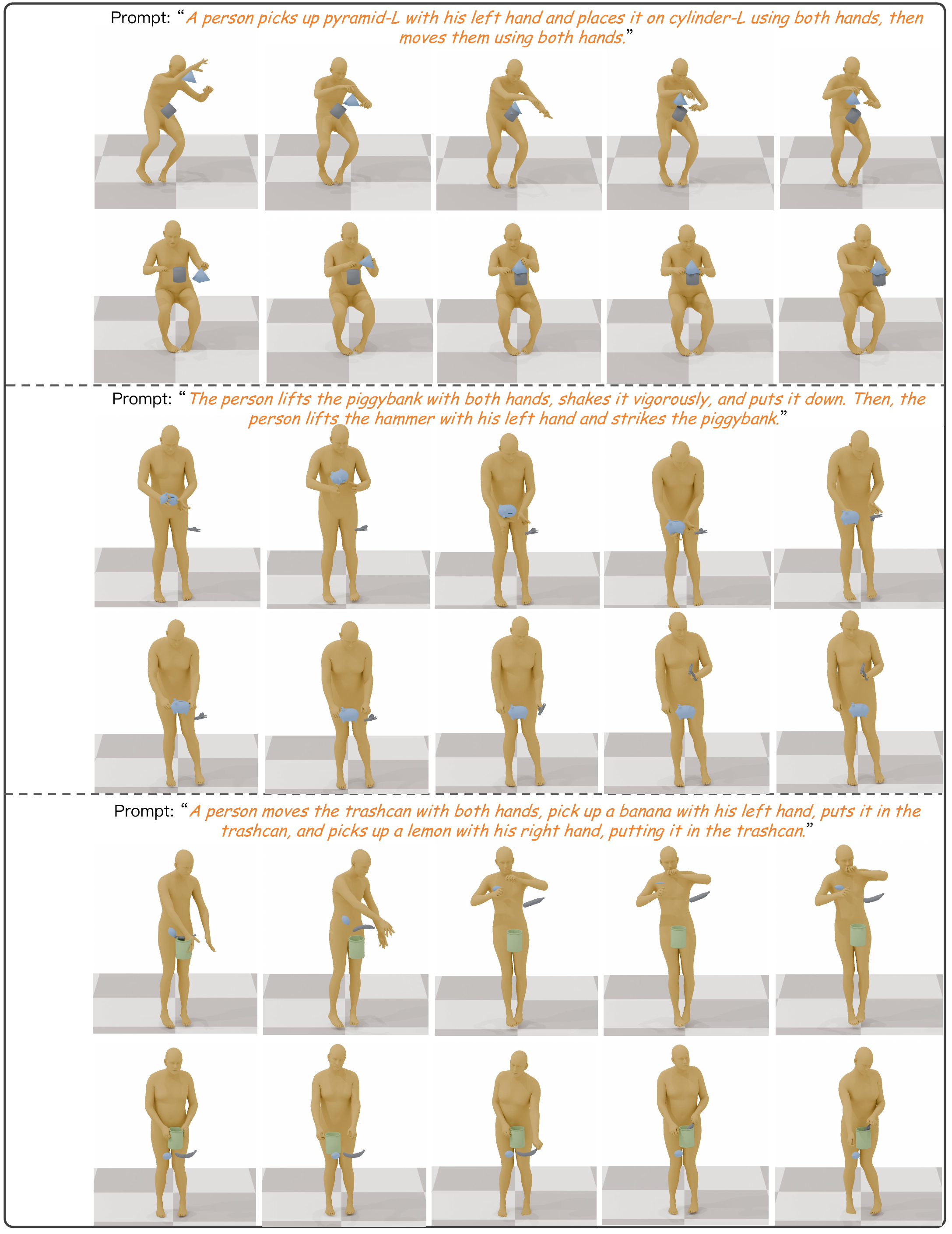}
     \put(10,30){\small{Ours}}
      \put(5,120){\small{HIMO-Gen}}
     \put(10,240){\small{Ours}}
      \put(5,320){\small{HIMO-Gen}}
      \put(10,440){\small{Ours}}
      \put(5,520){\small{HIMO-Gen}}
 
     \end{overpic}
    \vspace{-1em}
    \caption{\textbf{Qualitative comparisons on the test set of HIMO dataset}. Compared to baseline method, our approach generates more realistic and coherent human–object interactions, featuring more accurate contact and fewer artifacts.}
    \label{fig:vis_results_himo}
    \vspace{-1em}
\end{figure*}

\noindent\textbf{Additional results of ablation study.}
We compare our full model with several variants to evaluate the contribution of each component. Without the three-stage design (Separate Stage), the generated object motion becomes unstable, leading to physically implausible interactions. Without contact optimization, the object motion can still exhibit plausible patterns, such as opening or closing the laptop, but the human motion often fails to establish accurate contact, especially in complex multi-object scenarios. This suggests that contact optimization plays an important role in refining fine-grained human-object interactions after the initial motion generation.

Furthermore, we provide quantitative results for the ablation study in \Tref{table:ablation_main}. The results show that removing key components consistently degrades performance, while our full model achieves better motion quality and interaction accuracy across the evaluated metrics. These findings demonstrate the effectiveness of the proposed staged generation pipeline and the contact-aware refinement strategy.

\noindent\textbf{Additional visual results on OMOMO.}
We also provide additional qualitative results on the OMOMO dataset for single-object interactions, as shown in \Fref{fig:more_vis_omomo}. 
These examples demonstrate that our method also work well on the single-object setting, producing natural whole-body motions with accurate spatial alignment and physically plausible contacts.

\noindent\textbf{Additional visual results on HIMO benchmark.}
% We provide additional quantitative results on the HIMO benchmark in \Tref{table:quant_himo}, evaluated using the official evaluator. Our method achieves strong and competitive performance across the reported metrics, demonstrating its effectiveness on this dataset. Notably, our model can naturally adapt to varying numbers of interacting objects within a unified framework, without requiring changes to the model architecture or training procedure. In contrast, HIMO-Gen requires separate training and inference pipelines for each object count, which limits its flexibility and scalability.
We present additional qualitative results in \Fref{fig:vis_results_himo}. These examples further illustrate the ability of our method to generate coherent human-object interactions across diverse scenarios and varying object configurations.

\begin{figure*}[t]
% \vspace{-2em}
    \centering
     \begin{overpic}[trim={0 0 0 0}, width=1.0\linewidth]{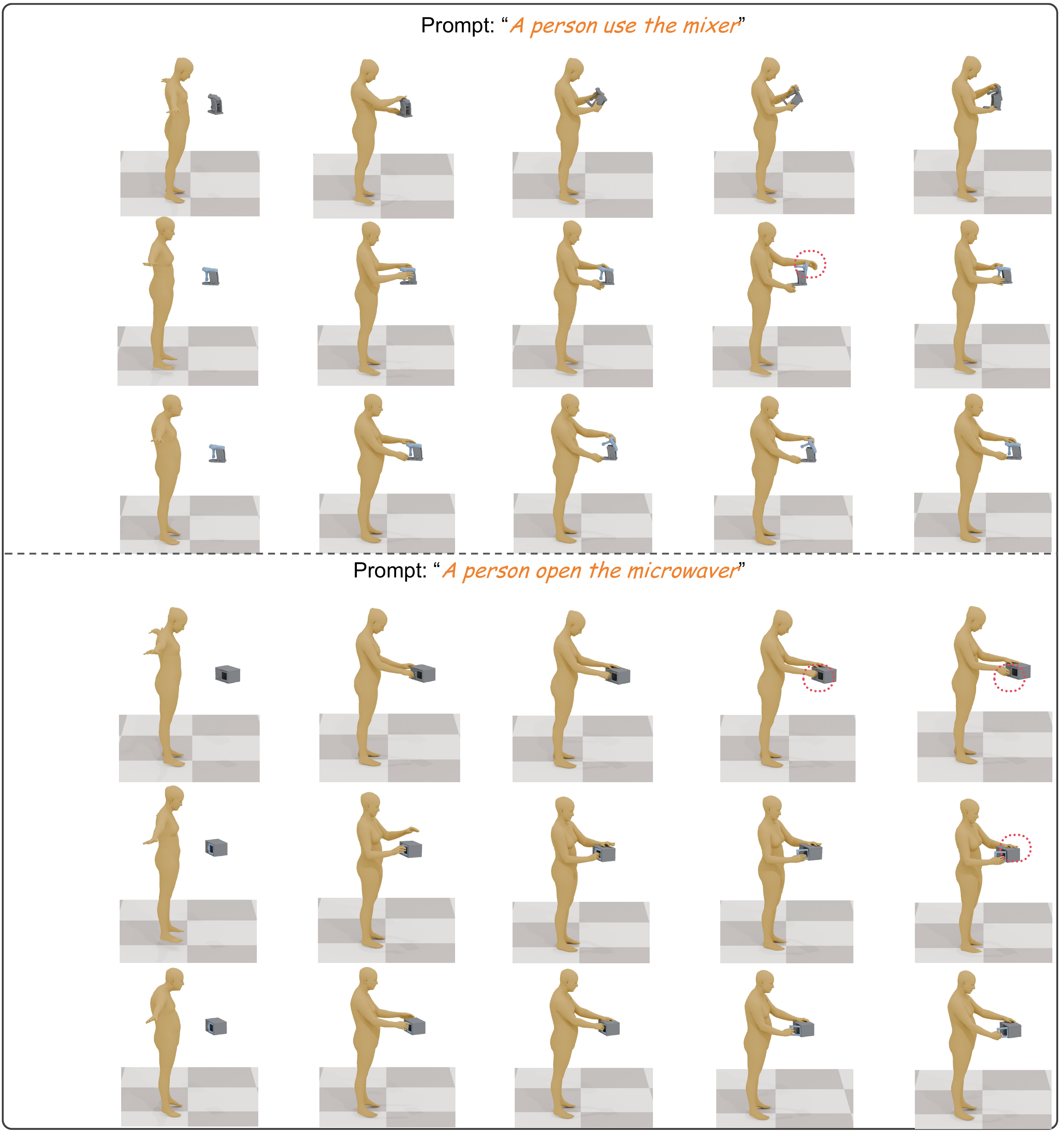}
     \put(10,30){\small{Ours}}
      \put(5,110){\small{HIMO-Gen}}
     \put(10,190){\small{CoDA}}
     \put(10,310){\small{Ours}}
      \put(5,390){\small{HIMO-Gen}}
      \put(10,470){\small{CoDA}}
     \end{overpic}
    \vspace{-1em}
    \caption{\textbf{Qualitative comparisons on the test set of ARCTIC dataset}. Compared to baseline methods, our approach generates more realistic and coherent human–object interactions, featuring more accurate contact and fewer artifacts. For HIMO-Gen, We modify its input and output layer to suit this articulated object motion. }
    \label{fig:vis_results_arctic}
    \vspace{-1em}
\end{figure*}

\noindent\textbf{Qualitative results on ARCTIC benchmark.}
% We provide additional quantitative results on the ARCTIC benchmark in \Tref{table:quant_arctic}, evaluated using the provided evaluator. Our method also achieves competitive performance on this benchmark.
We present qualitative results in \Fref{fig:vis_results_arctic}, which further demonstrate the effectiveness of our method in generating realistic and coherent interactions for articulated objects.

\noindent\textbf{Quantitative results on OMOMO benchmark using GT object motion.}
We report quantitative comparisons on the OMOMO dataset, where all methods are conditioned on the GT object motion. As a reference upper bound, we include GT-OMOMO~\cite{li2023object}, which utilizes ground-truth object motion as input.

We evaluate human motion quality using FID, R-Precision (R$_{prec}$), and Foot Sliding (FS), and measure the deviation from ground-truth motion using GT Difference metrics, including MPJPE (mean per-joint position error) and root translation error (T$_{root}$). These metrics quantify how closely the generated motion matches the ground-truth human motion in both joint positions and global translation.

As shown in ~\Tref{table:quant_cond_gt_obj}, our method achieves the best overall performance across most metrics. In particular, it yields the lowest FID and GT Difference errors (MPJPE and T$_{root}$), indicating improved motion realism and closer alignment with ground-truth motion. Compared to the variant without noise optimization, our full model further reduces reconstruction errors, demonstrating the effectiveness of the proposed optimization strategy.

\begin{table}[t]
\renewcommand{\arraystretch}{1.2}
\begin{center}
    \resizebox{1.0\linewidth}{!}{\begin{tabular}{l |aaa cc}
       \toprule
        & \multicolumn{5}{c}{\cellcolor{SceneOrange}\text{OMOMO Dataset}}  \\

            Method& \multicolumn{3}{a}{Human Motion} & \multicolumn{2}{c}{GT Difference} \\
          \ &  FID $\downarrow$  & $R_{prec}$ $\uparrow$  & FS $\downarrow$  & MPJPE $\downarrow$ & $T_{root}$$\downarrow$ \\
        \midrule
         % & \multicolumn{8}{c}{\cellcolor{LightGreen}\textit{Single Rigid Object}}  \\
         Real              &  0.00  & 0.566    & 0.0002 & -  &  -  \\ \midrule

         GT-OMOMO~\cite{li2023object}    & 6.79  & 0.492 & 0.0025   & 14.37  & 21.55 \\
         \midrule
         w$/$o Noise Optimization            & \underline{2.85}    & \underline{0.524}   & \textbf{0.0003}  &  \underline{12.78} & \underline{20.65}   \\
           \textbf{Ours}     & \textbf{2.78} &\textbf{0.531} & \textbf{0.0003} & \textbf{11.59} & \textbf{19.57} \\
    \bottomrule
\end{tabular}}
\vspace{-0.4mm}
\caption{\textbf{Quantitative results on the OMOMO dataset for conditioning on the GT object motion.} We compare our full method with existing baseline. The best results are highlighted in \textbf{bold}, and the second-best results are \underline{underlined.}}
\label{table:quant_cond_gt_obj}
\vspace{-2em}
\end{center}
\end{table}

\noindent\textbf{Quantitative results on the ParaHome for unseen multi-object composition.}
Test sequences contain object combinations never seen together during training, while every
individual object is observed in training. We report results over 118 two-object and 35
three-object held-out sequences as shown in ~\Tref{tab:quant_comp}. $R_{prec}$ is omitted because ParaHome uses templated captions,
causing heavy caption collisions under this split; interaction metrics are undefined for ground truth.
Under unseen compositions HIMO-Gen's object motion degrades sharply---foot sliding (FS) and object
jerk (Jerk$_{obj}$) are an order of magnitude higher---while our staged design keeps object motion
smooth (low FS/Jerk$_{obj}$), attains the lowest FID, and yields markedly higher contact accuracy;
the only regression is three-object penetration.

\begin{table}[t]
\centering
\small
\setlength{\tabcolsep}{6pt}
\begin{tabular}{l cc cc}
\toprule
& & & \multicolumn{2}{c}{Revolute} \\
\cmidrule(lr){4-5}
Method & Prismatic & Screw & ParaHome & ARCTIC \\
\midrule
HIMO-Gen            & 0.0041 & 0.3532 & 0.1589 & 0.2173 \\
CoDA                & N/A    & N/A    & $0.0000^{\dagger}$ & $0.0000^{\dagger}$ \\
Ours                & 0.0014 & 0.2493 & 0.0641 & 0.0157 \\
Ours\,+\,Correction & \textbf{0.0000} & \textbf{0.0307} & \textbf{0.0216} & \textbf{0.0011} \\
\bottomrule
\end{tabular}
\vspace{2pt}
{\footnotesize $\dagger$ Trivially zero: CoDA assumes a fixed, hardcoded hinge axis, so its
revolute residual is zero by construction and not directly comparable.}
\caption{Hinge compliance of generated articulated objects: deviation of generated parts from
their hinge axes (lower is better). We outperform HIMO-Gen on every joint type. CoDA attains $0$
on revolute joints only because it hardcodes a single hinge axis ($\dagger$), which also prevents
it from representing prismatic or screw joints (N/A); our surface-keypoint representation supports
all three. ``Ours\,+\,Correction'' projects the generated motion onto the constrained trajectory
given a predefined pivot/axis and joint type (no retraining), leaving only a small residual while
keeping generation joint-type-agnostic.}
\label{tab:hinge_compliance}
\end{table}

\noindent\textbf{Hinge compliance of generated articulated objects.}
We measure hinge compliance as the deviation of generated articulated parts from their hinge axes
(lower is better; \Tref{tab:hinge_compliance}). Our method outperforms HIMO-Gen across all
joint types. CoDA attains zero deviation on revolute joints only because it hardcodes a single
hinge axis, which also prevents it from representing prismatic or screw joints at all; in contrast,
our surface-keypoint representation supports all three joint types without any per-mechanism
assumption. Finally, ``Ours\,+\,Correction'' takes the generated motion together with a predefined
pivot/axis and articulation type and projects it onto the corresponding constrained trajectory
(no retraining), substantially reducing the residual while keeping the generator
joint-type-agnostic. This is a favorable trade-off: CoDA achieves exact compliance for a single
mechanism, whereas we support all three with only a small, correctable residual.

\noindent\textbf{Component-slot ordering and permutation sensitivity.}
Our model represents an interaction as a set of per-component token streams
(Sec.~3.2), so the assignment of physical components to input slots is arbitrary.
During preprocessing on ParaHome we randomize this ordering, so the model is
exposed to varied slot assignments throughout training, and the object tokens are
processed jointly by the Transformer rather than in any fixed per-slot channel.
To verify that generation is insensitive to this choice, we re-evaluate the
two-object ParaHome split under the original component order and three random slot
permutations. As reported in ~\Tref{tab:perm_sensitivity}, FID and both
contact-accuracy metrics are essentially unchanged---FID varies by only $0.02$ and
the contact accuracies by $0.002$ across permutations---confirming that our results
do not depend on how components are assigned to slots.

\begin{table}[h]
\centering
\small
\setlength{\tabcolsep}{6pt}
\begin{tabular}{ll r}
\toprule
Modality & Metric & Value \\
\midrule
Human motion  & MPJPE (mm)$\downarrow$                    & 27.52 \\
Object motion & Vertex error (mm)$\downarrow$              & 16.28 \\
\midrule
\multirow{4}{*}{Contact field}
              & MAE$\downarrow$                            & 0.0073 \\
              & RMSE$\downarrow$                           & 0.0205 \\
              & MAE on contact cells ($>\!0.5$)$\downarrow$ & 0.0665 \\
              & Precision / Recall / F1$\uparrow$          & 0.928 / 0.927 / 0.927 \\
\bottomrule
\end{tabular}
\caption{Reconstruction quality of the frozen Causal Autoencoder on held-out ParaHome/OMOMO
sequences. Because all three stages generate in this latent space (Sec.~\ref{sec:supp_ae}), these
numbers upper-bound the fidelity attainable by the diffusion models. Motion errors are
per-marker/per-vertex $L_2$ in millimeters; contact-field errors are on the $[0,1]$ distance field,
and precision/recall/F1 use a $0.5$ threshold.}
\label{tab:ae_recon}
\end{table}

\noindent\textbf{Autoencoder reconstruction quality.}
Because every stage generates in the latent space of the frozen autoencoder
(Sec.~\ref{sec:supp_ae}), its reconstruction quality upper-bounds the fidelity of the full
pipeline. ~\Tref{tab:ae_recon} reports encode--decode error on held-out sequences. Human motion
is recovered with a mean per-marker error of $27.5$\,mm and object keypoints with $16.3$\,mm, both
small relative to body and object scale. The contact distance field is reconstructed almost exactly
(MAE $0.007$ over the $[0,1]$ field); although the field is dominated by a near-zero background, the
per-marker contact codec recovers the actual contacts with well-balanced precision and recall
($0.928/0.927$; F1 $0.927$ at a $0.5$ threshold), confirming that the latent space preserves the
fine-grained, sparse contact structure the downstream stages depend on.

\begin{table}[h]
\centering
\small
\begin{tabular}{lcc}
\toprule
Metric (mm)$\downarrow$ & Regression only & +\,Test-time Optimization \\
\midrule
Marker error & 101.83 & \textbf{21.88} \\
Joint MPJPE  & 99.96  & \textbf{33.72} \\
Trans error  & 60.32  & \textbf{18.93} \\
\bottomrule
\end{tabular}
\caption{Marker-to-SMPL-X fitting error (mm) on held-out sequences, before and after the
400-step test-time optimization.}
\label{tab:smplx_fit}
\end{table}

\noindent\textbf{Quantitative kinematic-validity measurements.}
Although articulated components are generated without explicit joint constraints, we quantify how
often the outputs are kinematically invalid by thresholding per-frame residuals, evaluated on
$282$ articulated part--base pairs across $222$ ParaHome sequences (\Tref{tab:kinematic_validity}).
\emph{Joint axis} measures the deviation of the per-frame relative-rotation axis between a part and
its base from the fitted hinge axis; it extends our hinge-compliance residual by thresholding the same per-frame residual at $10^\circ$ to yield a violation \emph{rate} rather than a mean deviation. \emph{Joint limits} flag frames whose recovered
joint value exceeds the per-category range measured from ground truth by more than $10\%$, verifying
that the motion magnitude stays physical. \emph{Assembly connectivity} fits a single shared
base-to-part pivot by least squares and reports its residual (connection drift), flagging frames
whose drift exceeds $2$\,cm; this verifies that the two parts remain assembled around one consistent
pivot. Joint limits and connectivity are complementary---the former checks motion magnitude, the
latter structural consistency. Across all three aspects the violation rates are low (frame-level
$\le\!5.2\%$, with a mean pivot drift of only $0.43$\,cm and a median of $0$), showing that the
generated articulated motions are kinematically valid the large majority of the time even without
hard constraints. The small residual violations are precisely what the optional projection step removes.

\noindent\textbf{Robustness of Kabsch pose recovery.}
We recover each component's per-frame $SE(3)$ pose from its predicted keypoints using the Kabsch
algorithm (SVD with reflection handling), which returns an exact $SO(3)$ rotation for any
non-collinear keypoint set. The canonical keypoints are selected by farthest-point sampling, which
maximizes their spread, so the triplets are well conditioned across all 87 objects (median
anisotropy $\sigma_2/\sigma_1{=}0.65$, mean minimum interior angle $42^\circ$). Recovery stays
stable while $\sigma_2/\sigma_1{>}0.2$ and degrades only as the keypoints approach collinearity
($\sigma_2/\sigma_1{<}0.1$), a regime reached only by thin, elongated objects (e.g., knife, pen),
which we state explicitly as a limitation. Under injected keypoint noise the rotation error grows
roughly linearly ($\approx\!1.3^\circ/$mm for $k{=}3$; ~\Tref{tab:kabsch}). Because our
formulation and solver accept any number of keypoints, an over-determined set ($k>3$) both improves
noise robustness and eliminates the collinearity failure mode, and we therefore adopt $k>3$ for
elongated objects.

\begin{table}[t]
\centering
\small
\setlength{\tabcolsep}{6pt}
\begin{tabular}{l l c}
\toprule
Validity aspect & Threshold & Frames$\downarrow$ \\
\midrule
Joint axis            & $>\!10^\circ$ axis deviation   & 1.46\%  \\
Joint limits          & $>$ GT range $+\,10\%$          & 0.02\%  \\
Assembly connectivity & $>\!2$\,cm pivot drift          & 5.20\%  \\
\bottomrule
\end{tabular}
\caption{Kinematic-validity violation rates on ParaHome (282 articulated part--base pairs across
222 sequences). Components are generated \emph{without} explicit joint constraints; we threshold
per-frame residuals to measure how often the outputs are kinematically invalid. A frame is a
violation if its residual exceeds the threshold; Assembly connectivity is reported at the frame level, with mean/median pivot drift of $0.43$/$0.00$\,cm. All rates are low.}
\label{tab:kinematic_validity}
\end{table}

\begin{table}[t]
\centering
\small
\setlength{\tabcolsep}{8pt}
\begin{tabular}{l ccc}
\toprule
Keypoint noise & $k{=}3$ & $k{=}6$ & $k{=}10$ \\
\midrule
5\,mm  & $6.3^\circ$  & $3.8^\circ$ & $2.9^\circ$ \\
10\,mm & $13.8^\circ$ & $6.5^\circ$       & $4.6^\circ$          \\
\bottomrule 
\end{tabular}
\caption{Robustness of Kabsch pose recovery. We apply a known $SE(3)$ to the canonical keypoints,
perturb them with isotropic Gaussian noise, and recover the pose by Kabsch (SVD with reflection
handling); we report rotation error (degrees). An over-determined keypoint set ($k>3$) markedly
improves robustness and removes the near-collinear failure mode. This i.i.d.\ setting is a worst
case---the model's per-frame keypoints are temporally correlated and smoother, and translation
recovery is considerably more stable.}
\label{tab:kabsch}
\end{table}

\noindent\textbf{Effect of the number of object keypoints.}
\begin{table}[h]
\renewcommand{\arraystretch}{1.2}
\begin{center}
    \resizebox{\linewidth}{!}{\begin{tabular}{|c |aaaaa ccc|}
       \hline
         % \cmidrule{2-13}
        % \hline
         Num. of  & \multicolumn{5}{a}{Motion} & \multicolumn{3}{c|}{Interaction} \\
          Keypoints\ &  FID $\downarrow$  & $R_{prec}$ $\uparrow$ & Div $\rightarrow$ & FS $\downarrow$ &Jerk$_{obj} \downarrow$ & $C_{acc}^{tem}  \uparrow$ & $C_{acc}^{body} \uparrow$  & Pene $\downarrow$ \\
        \hline
         3 & \textbf{6.09}  & \textbf{0.598}  & \textbf{7.57}   &\textbf{ 0.0017} & 0.40 &\textbf{ 0.680}  & \textbf{0.906} & 0.776       \\
         6 & 6.21  & 0.576  & 7.81   & 0.0021  & 0.42  & 0.673  & 0.894 & 0.721      \\
         9 & 6.18  & 0.582  & 7.74   & 0.0020  & \textbf{0.38}  & 0.668  & 0.890 & 0.716  \\
        \hline
\end{tabular}}
\caption{\textbf{Quantitative results on ParaHome (3 objects) with different numbers of keypoints $K$.} Best results are highlighted in bold.}

\label{table:number_keypoints}
\end{center}
\end{table}
\Tref{table:number_keypoints} studies how the number of sampled object keypoints $K$ affects generation quality and interaction plausibility.
Overall, we observe a clear trade-off between motion fidelity and interaction accuracy.
Using fewer keypoints ($K{=}3$) yields the best overall performance, achieving the lowest FID, \textit{R}$_{prec}$, and foot sliding score (FS), which indicates smoother and consistent motions. It also provides the strongest interaction quality, with the highest contact accuracies. 
In contrast, increasing the number of keypoints does not further improve interaction plausibility and instead degrades object human motion quality (e.g.,~\textit{R}$_{prec}$), and contact accuracy, suggesting that overly dense object representations may overfit local geometry and introduce unnecessary constraints during generation. 
Based on these results, we set $K{=}3$ as a good trade-off between motion realism and physically plausible interactions.

\begin{table*}[t]
\renewcommand{\arraystretch}{1.0}
\begin{center}
    \resizebox{1.0\linewidth}{!}{\begin{tabular}{l |aaaa ccc| aaaa ccc}
       \toprule
        & \multicolumn{7}{c|}{\text{ParaHome — Unseen Compositions (2 Objects)}} & \multicolumn{7}{c}{\text{ParaHome — Unseen Compositions (3 Objects)}} \\
            Method& \multicolumn{4}{a}{Motion} & \multicolumn{3}{c|}{Interaction} &\multicolumn{4}{a}{Motion} & \multicolumn{3}{c}{Interaction}\\
          \ &  FID $\downarrow$  & Div $\rightarrow$ & FS $\downarrow$ & Jerk$_{obj}$ $\downarrow$ & $C_{acc}^{tem}\uparrow$ & $C_{acc}^{body}$$\uparrow$ & Pene $\downarrow$  &  FID $\downarrow$  & Div $\rightarrow$ & FS $\downarrow$ & Jerk$_{obj}$ $\downarrow$ & $C_{acc}^{tem}\uparrow$ & $C_{acc}^{body}$$\uparrow$ & Pene $\downarrow$ \\
        \midrule
          & \multicolumn{14}{c}{\cellcolor{LightGreen}\textit{Unseen Multi-Object Compositions}}  \\
           Real              &  0.00  & 6.18   & 0.0059 & 0.14 & - &  -    & -    & 0.00    & 5.00  & 0.0091  & 0.14     & -     &  -   &- \\

          HIMO-Gen              &  12.37  &7.63 & 0.2791 & 4.89 & 0.511 &  0.623    & 0.622    & 10.87     & 4.52  & 0.3271  & 6.28     & 0.543     &  0.627   & 0.578 \\

           Ours              &  7.81  & 6.96 & 0.0132 & 0.59 & 0.692 &  0.863    & 0.516    & 8.32     & 5.82  & 0.0571  & 0.48     & 0.675     &  0.821   & 0.663 \\
        \bottomrule
\end{tabular}}
\caption{\textbf{Ground-truth (Real) statistics on the ParaHome \emph{unseen multi-object composition} split.} Test sequences contain object combinations never seen together during training (e.g.\ \texttt{\{book, bookshelf, desk\}}, \texttt{\{cutting board, pan\}}), while every individual object is observed in training. We report GT reference statistics over 118 (2-object) and 35 (3-object) held-out sequences. $R_{prec}$ / Matching Score are omitted here: ParaHome uses templated captions, so restricting to a few held-out compositions causes heavy caption collisions (e.g.\ 41 sequences share one caption), which makes batch-wise text-to-motion retrieval degenerate; interaction metrics are undefined for ground truth.}
\label{tab:quant_comp}
\end{center}
\end{table*}

\noindent\textbf{Effect of contact representations.}
We compare three types of intermediate contact supervision: our \emph{distance field}, a \emph{binary contact label}, and the raw \emph{Euclidean distance}. 
For evaluation, we directly threshold the predicted contact signals to obtain binary contact labels and compare them against the ground-truth labels using the contact metrics described in Sec.~\ref{supp:metrics}.

As shown in \Tref{table:contact_type}, the proposed distance field achieves the best overall results across all metrics, indicating that it provides a more informative and learnable interaction cue.

In contrast, the binary label leads to a clear performance drop. 
We attribute this to its extreme sparsity: most marker--object pairs are non-contact and thus take zero values, resulting in weak gradients and making it difficult for the model to learn fine-grained correspondence and contact switching over time. 
Using raw Euclidean distances, as in ROG~\cite{xue2025guiding}, also degrades performance, as they are absolute and unnormalized measures with a large dynamic range.
This increases the learning complexity and makes training sensitive to scale variations across different objects and motions. 

% Overall, our distance-field formulation strikes a better balance between informativeness and numerical stability, leading to more accurate and physically plausible HOI generation.
Some works utilize Signed Distance Fields (SDF) to encode object geometry, which represents signed inside/outside information. While SDF is a powerful representation for geometry modeling, we find that it is not necessary for our interaction-centric objective. Our goal is to model interaction-aware proximity between human and object surfaces over time. Overall, our distance-field formulation strikes a better balance between informativeness and numerical stability, leading to more accurate and physically plausible HOI generation.
% \begin{figure}[h]
%     \centering
%     \begin{overpic}[width=0.92\linewidth]{figures/questionnaire.pdf}
%     \end{overpic}
%     % \vspace{-1.6em}
% \caption{\textbf{Illustration of questionnaire in our user study.} }
% % \vspace{-1em}
%     \label{fig:questionnaire}
% \end{figure}

\begin{figure}
    \centering
    \includegraphics[width=1.0\linewidth]{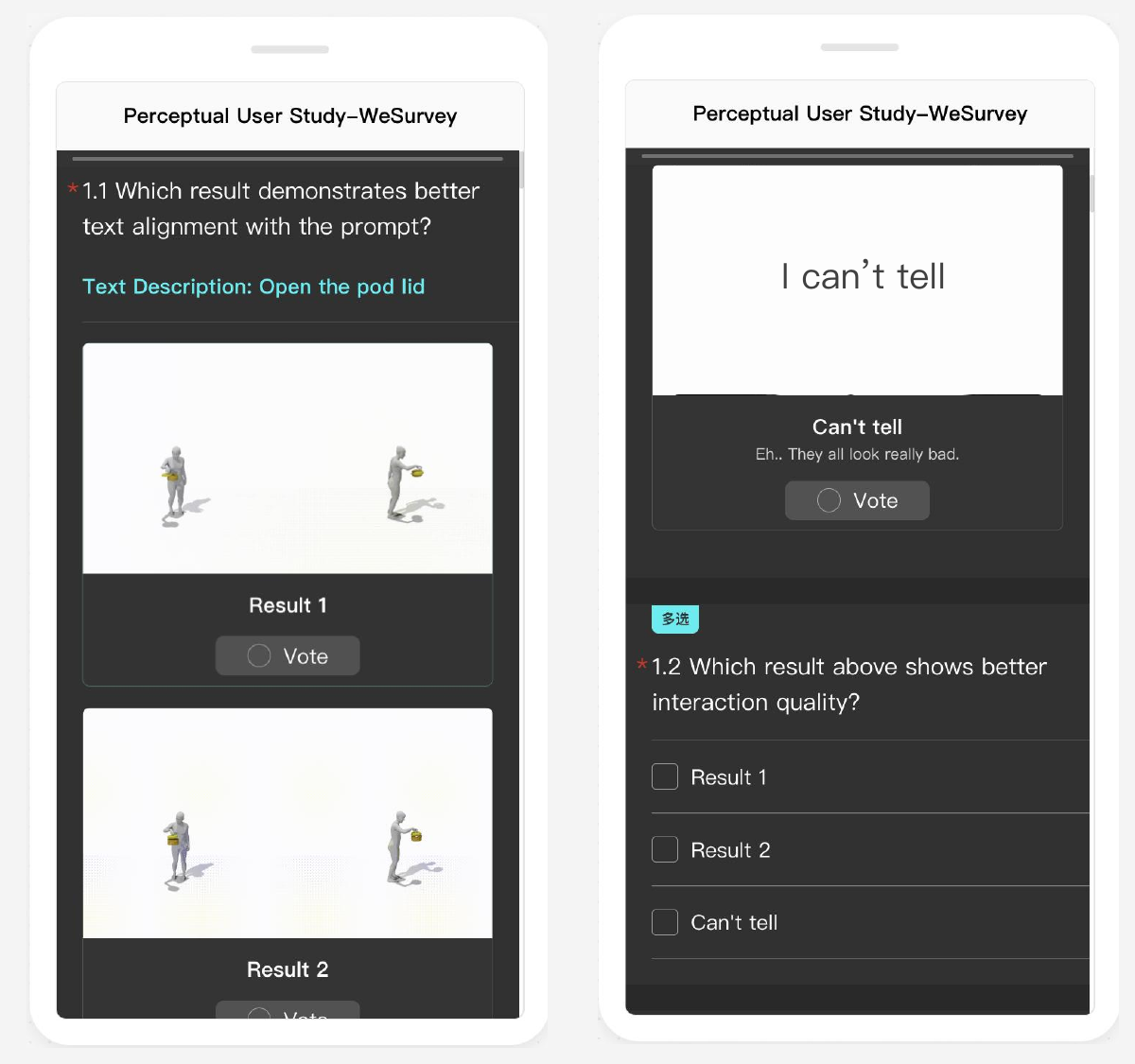}
\caption{\textbf{Illustration of questionnaire in our user study.} }
    \label{fig:questionnaire}
\end{figure}

\noindent\textbf{Inference speed.}
\Tref{table:inference_time} report the inference time of each module in our pipeline. All timings are measured on a single NVIDIA A6000 GPU for generating a 124-frame sequence. The overall runtime is primarily dominated by the contact optimization in Stage III, which involves iterative ODE-based sampling and gradient-based updates. Nevertheless, our method remains significantly more efficient than CoDA~\cite{pi2025coda}. In particular, CoDA does not support batch-level optimization and requires approximately 14 minutes per sample, whereas our approach enables more efficient batched processing.

\begin{table}[t]
\centering
\small
\setlength{\tabcolsep}{6pt}
\begin{tabular}{l ccc}
\toprule
Permutation & FID$\downarrow$ & $C^{tem}_{acc}\uparrow$ & $C^{body}_{acc}\uparrow$ \\
\midrule
Original order    & 4.49 & 0.669 & 0.896 \\
Random perm.\ \#1 & 4.52 & 0.665 & 0.893 \\
Random perm.\ \#2 & 4.47 & 0.671 & 0.898 \\
Random perm.\ \#3 & 4.51 & 0.667 & 0.895 \\
\midrule
Mean\,$\pm$\,std  & $4.50\pm0.02$ & $0.668\pm0.002$ & $0.895\pm0.002$ \\
\bottomrule
\end{tabular}
\caption{Sensitivity to component-slot ordering on the ParaHome two-object split. We re-run
generation with the input components assigned to different slot orders. FID and contact accuracy
are stable across permutations (std $\le\!0.02$), showing our model is effectively invariant to
component ordering.}
\label{tab:perm_sensitivity}
\end{table}

% \section{Additional Qualitative Results.}
% \label{supp:qualitative}

\begin{table}
    \centering
    \vspace{2pt}
    \begin{tabular}{lc}
        \toprule
        \textbf{Module} & \textbf{Time} \\
        \midrule
        Object Motion Generation & 0.22 secs \\
        Contact Distance Field Prediction & 0.59 secs \\
        Body Motion Synthesis w/ Contact Optimization & 4.21 mins \\
        \bottomrule
    \end{tabular}
    \caption{\textbf{Inference time.}}
    \label{table:inference_time}
    \vspace{-6pt}
\end{table}

\section{Details of User Study}
\label{supp:user_study}

\Fref{fig:questionnaire} illustrates the questionnaire interface. For each trial, participants are shown the text description and a set of anonymized generated animations. For single rigid object interactions, we run a three-way comparison among our method, CHOIS, and HOI-Diff. For multiple/articulated objects interactions, we run a pairwise comparison between our method and HIMO. Participants provide two judgments per trial: \textbf{text alignment} (which animation best matches the prompt) and \textbf{interaction quality} (which looks more natural and physically plausible), with an optional \textbf{“Can’t tell”} choice when differences are unclear. To minimize presentation bias, we randomize the placement of results across trials. For fairness, we keep the camera consistent across methods. 

\section{Failure Cases}
\label{supp:limitations}

\begin{figure}[t]
% \vspace{-2em}
    \centering
     \begin{overpic}[trim={0 0 0 0}, width=0.98\linewidth]{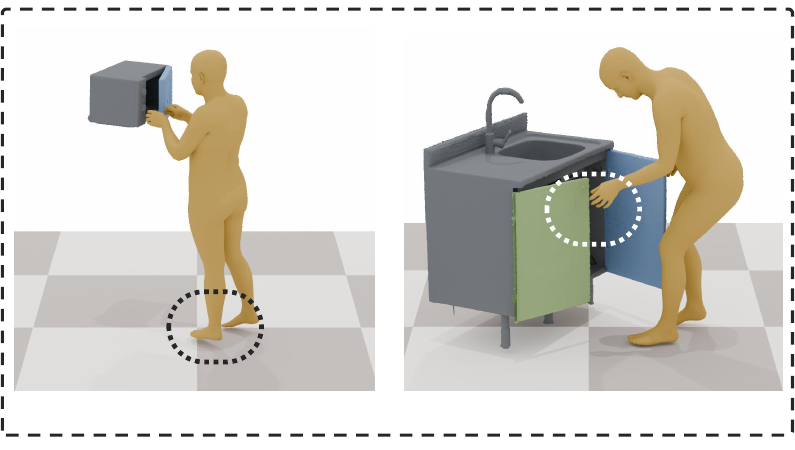}
     \put(38,9){\small{foot floating}}
      \put(158,9){\small{contact missing}}
 
     \end{overpic}
    \vspace{-1em}
    \caption{\textbf{Representative failure cases.} Left: errors in body fitting lead to incorrect body scale and foot floating. Right: temporally inconsistent distance field predictions result in missed contacts.}
    \label{fig:failure_cases}
    \vspace{-1em}
\end{figure}

% \noindent\textbf{Failure cases.} 
In \Fref{fig:failure_cases}, we present several failure cases of our method. First, the fitting model is not always accurate; in some instances, it recovers a human mesh with a smaller height, leading to foot floating, as shown on the left. Additionally, contact may be missed when the predicted distance field is not sufficiently accurate in the temporal dimension.

% \noindent\textbf{Limitations.} 
% First, our method relies on a noise optimization step like DNO~\cite{karunratanakul2024optimizing} for contact refinement, which is computationally more expensive than purely feed-forward generative approaches and therefore limits real-time deployment as shown in the \Tref{table:inference_time}. 
% Second, our current training and evaluation are conducted on ParaHome and OMOMO, whose object categories and geometric diversity remain limited. As a result, the model may not generalize robustly to novel objects with substantially different shapes or mechanisms without additional data or adaptation. 
% Finally, the interaction patterns in these datasets are dominated by hand-centric manipulation. Consequently, our model currently observes relatively few examples involving richer full-body contacts (e.g., sitting, kicking with the torso or foots), which constrains its ability to synthesize diverse body-part interactions. 
% Addressing these limitations will require larger-scale datasets with broader object coverage and more varied whole-body interaction types, as well as more efficient optimization or fully amortized refinement strategies.

\section{Supplemental Video.}
\label{supp:video}
We provide a supplemental video to qualitatively demonstrate the effectiveness of our method. 
The video includes: (i) background and motivation; (ii) additional comparisons with baseline methods on ParaHome and OMOMO; 
(iii) diverse examples of multi-object manipulation and articulated object interactions; 
(iv) ablation results illustrating the contributions of key components; (v) predicted contact distance field; and (vi) failure cases and more our visual results. 
We recommend viewing the video in full screen to better inspect contact events and interaction details.

\end{document}